%% file: main.tex
\documentclass{article}
\usepackage{iclr2027_conference,times}
\usepackage{amsmath,amssymb,amsthm,bm}
\usepackage{fix-cm}
\usepackage{booktabs,array,longtable,tabularx,graphicx,multirow}
\usepackage{tikz,pgfplots}
\usetikzlibrary{arrows.meta,positioning,calc,fit}
\usepgfplotslibrary{groupplots,fillbetween}
\usepackage[T1]{fontenc}
\usepackage{tgheros}
\input{figures/style}
\pgfplotsset{compat=1.18}
\usepackage{microtype}
\usepackage[hidelinks,hypertexnames=false]{hyperref}
\usepackage{url}
\hypersetup{pdftitle={HiPPO Functional Memory: Coordinate-Queryable Attention for Irregular Time Series},pdfauthor={Zhaodong Guo and Haoyun Feng}}
\usepackage{enumitem}
\usepackage{placeins}
\usepackage{wrapfig}
\usepackage{colortbl}
\definecolor{rowours}{HTML}{EAF1FB}
\definecolor{grouprow}{HTML}{F1F1F1}
\usepackage{float}
\usepackage{needspace}
\usepackage{algorithm}
\usepackage{algpseudocode}
\newtheorem{proposition}{Proposition}[section]
\newtheorem{theorem}[proposition]{Theorem}

\theoremstyle{definition}

\theoremstyle{definition}

\newcommand{\FM}{\operatorname{FM}}

\newcommand{\spanof}{\operatorname{span}}

\newcommand{\R}{\mathbb R}
\newcommand{\norm}[1]{\left\lVert#1\right\rVert}
\newcommand{\smallpm}{\,\mathord\pm\,}
\title{HiPPO Functional Memory: Coordinate-Queryable Attention for Irregular Time Series}
\author{\makebox[\dimexpr\textwidth-2\tabcolsep\relax][c]{Zhaodong Guo$^{1}$\qquad Haoyun Feng$^{1}$}\\
\makebox[\dimexpr\textwidth-2\tabcolsep\relax][c]{\small $^{1}$Imperial College London}}
\begin{document}
\maketitle
\begin{abstract}
\input{sections/abstract}
\end{abstract}
\input{sections/introduction}
\input{sections/related_work}
\input{sections/method}
\FloatBarrier
\input{sections/experiments}
\input{sections/conclusion}
\label{main-text-end}
\FloatBarrier
\input{sections/statements}
\bibliography{references}
\bibliographystyle{iclr2027_conference}
\clearpage
\appendix
\input{appendix/method_details}

\input{appendix/theory}
\input{appendix/protocols}
\input{appendix/additional}

\input{appendix/diagnostics}
\end{document}

%% file: figures/style.tex
\definecolor{fmBlue}{HTML}{0072B2}
\definecolor{fmOrange}{HTML}{D55E00}
\definecolor{fmTeal}{HTML}{00866A}
\definecolor{fmPurple}{HTML}{8064A2}
\definecolor{fmInk}{HTML}{202B3B}
\definecolor{fmGray}{HTML}{677586}
\definecolor{fmRule}{HTML}{D8E0E8}
\definecolor{fmPaper}{HTML}{F8FAFC}
\newcommand{\figfont}{\sffamily\fontsize{8.5}{10}\selectfont}
\newcommand{\figsmall}{\sffamily\fontsize{7.5}{9}\selectfont}
\newcommand{\figtitle}{\sffamily\bfseries\fontsize{9.5}{11}\selectfont}
\tikzset{
 fmArrow/.style={-{Latex[length=1.65mm,width=1.15mm]},line width=.8pt,draw=fmGray},
 fmBox/.style={rounded corners=2.5pt,line width=.65pt,align=center,inner sep=4pt,text=fmInk},
 fmLabel/.style={font=\figfont,text=fmInk,align=center},
 fmNote/.style={font=\figsmall,text=fmGray,align=center}
}
\pgfplotsset{
 fmAxis/.style={
 font=\figfont,axis lines*=left,axis line style={fmGray,line width=.55pt},
 tick align=outside,tick style={fmGray,line width=.5pt},
 tick label style={font=\figsmall,text=fmInk},
 label style={font=\figfont,text=fmInk},
 title style={font=\figtitle,text=fmInk},
 legend style={font=\figsmall,text=fmInk,draw=none,fill=white,inner sep=2pt},
 legend cell align=left,minor tick num=0,
 grid=none,scaled ticks=false,clip=false,
 /pgf/number format/1000 sep={\,}
 }
}

%% file: sections/abstract.tex
Irregular time series are recorded at nonuniform times and are often queried at other times: between observations, after the last one, or on a regular grid. Continuous-time models answer such queries by integrating latent dynamics, and time-aware attention models answer them by revisiting the observation tokens. We introduce HiPPO Functional Memory (HiPPO-FM), an attention operator that writes an irregular history into a rank-$M$ function space with explicit time coordinates. Masked self-attention first exchanges content among the observations; the updated values are then contracted against kernel-smoothed HiPPO--Legendre features and solved through a regularized Gram matrix, so the retained coefficients define a function that is read at any time through the same features. The write and read have the form of an interdomain sparse Gaussian-process mean: every readout lies in an $M$-dimensional reproducing-kernel subspace, and a pointwise bound separates the approximation error of this subspace from the error of the learned memory. Removing the functional memory from the encoder raises Spiral reconstruction RMSE 18-fold. The full model has the lowest error in 11 of 12 Spiral settings, the highest overall accuracy and best average rank on 20 irregular UEA datasets, the highest event-type accuracy on all five multi-type event benchmarks, and a 21\% lower extrapolation MSE than CRU on PhysioNet. Because the memory has fixed rank, the state retained after encoding keeps a fixed size as the history grows and is $102\times$ smaller than a token cache at 1,024 observations.

%% file: sections/introduction.tex
\section{Introduction}
\label{sec:intro}

Clinical records, event-triggered sensors, and user activity logs are observed at data-dependent times, while the quantities of interest are requested at task-dependent times: between two measurements, on a regular grid, or after the last measurement. A model for such data must decide how the history is stored, because that representation determines how an unobserved time is answered.

Most methods store the history in one of two forms \citep{rubanova2019latent,kidger2020neural,schirmer2022cru,shukla2021mtan,chen2023contiformer}. Latent ODEs, Neural CDEs, and CRU keep a latent state and answer a requested time by integrating it to that time, which requires sequential solver steps whose number grows with the horizon. mTAN and ContiFormer keep the observation tokens and answer a requested time by attending over them, so the representation stays indexed by the observations. Patch-, graph-, and point-based forecasters reorganize the observations before prediction \citep{zhang2023warpformer,yalavarthi2024grafiti,zhang2024tpatchgnn,hu2025vimts,liu2026apn,liu2026astgi}. In each case the stored representation has coordinates in observation index or solver time, and a requested time is related to it only through further computation.

HiPPO \citep{gu2020hippo} offers a different representation: a signal history is projected onto a finite orthogonal-polynomial basis, and the coefficients describe the history as a function of time. Interdomain Gaussian processes \citep{lazaro2009interdomain} show how such projections pair with a kernel: their cross-covariance with the observations defines a write, and the inverse Gram matrix of the projections defines a read whose output lies in a fixed function space. OHSVGP uses this pair with HiPPO projections for online sparse variational inference \citep{NEURIPS2025_c4025018}. We use the same pair as the memory of an attention layer, so that the memory is a function rather than a set of tokens.

\begin{figure}[t]
\centering
\includegraphics[width=\linewidth]{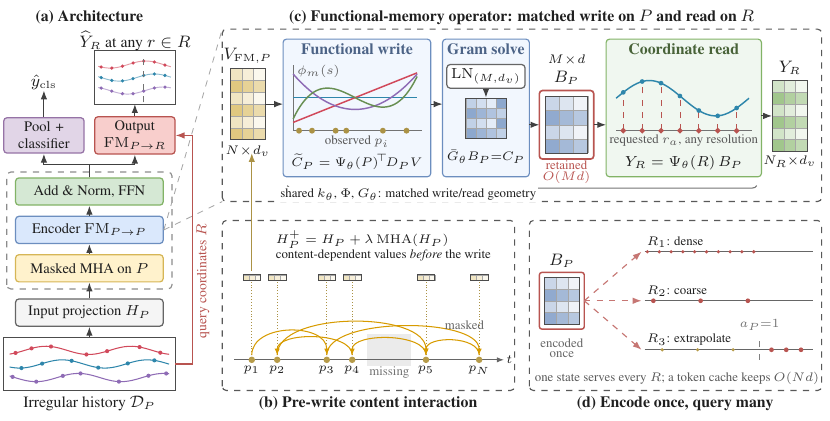}
\caption{Overview of HiPPO Functional Memory (HiPPO-FM).
\textbf{(a) Architecture.} An irregular history is projected, passed through masked self-attention on the observed support and an encoder FM operator, and either pooled for sequence-level outputs (UEA, events) or read by an output FM head at requested coordinates $R$.
\textbf{(b) Pre-write content interaction.} Self-attention over the irregular observed support forms content-dependent values before the finite-rank write.
\textbf{(c) FM operator.} Values are contracted against kernel-smoothed Legendre features at the observed coordinates, normalized, and solved through the regularized Gram matrix into the retained coefficients $B_P\in\mathbb R^{M\times d}$; the same kernel, basis, and Gram matrix evaluate $B_P$ at the requested coordinates.
\textbf{(d) Reuse.} One encoded state serves dense, coarse, and extrapolated query sets.}
\label{fig:hippo_fm}
\end{figure}

The resulting operator, \textbf{HiPPO Functional Memory (HiPPO-FM)}, compiles an observed support $P$ into retained coefficients $B_P\in\mathbb R^{M\times d}$ and evaluates them at requested coordinates $R$,
\[
\mathcal D_P \longmapsto B_P,
\qquad O(R)=\Psi_\theta(R)B_P,
\]
where $\Psi_\theta$ are kernel-smoothed shifted-Legendre features. The write weights depend only on the observation times, so interaction between observation contents has to happen before the write: HiPPO-FM applies masked self-attention on the observed support and then writes the updated values (Figure~\ref{fig:hippo_fm}). The memory thus combines content-dependent values with a time-coordinate system shared by writing and reading.

The functional bottleneck carries the accuracy of the model: removing it from the encoder while keeping self-attention raises Spiral RMSE from $0.54$ to $9.88\times10^{-2}$ in all five seeds. With it, HiPPO-FM has the lowest error in 11 of 12 Spiral settings, the highest overall accuracy and best average rank on 20 irregular UEA datasets, the highest event-type accuracy on all five multi-type event benchmarks, and a 21\% lower extrapolation MSE than CRU on PhysioNet. The fixed rank also keeps the retained state at a constant size, $102\times$ smaller than a token cache at $N=1024$, and a training iteration costs 29--776$\times$ less than the ODE, CDE, and ContiFormer baselines.

Our contributions are:
\begin{itemize}[leftmargin=1.3em,itemsep=1pt,topsep=2pt]
\item \textbf{Operator.} HiPPO-FM writes self-attended values into a fixed-rank memory of kernel-smoothed HiPPO--Legendre functionals and reads it at arbitrary times through the matched inverse-Gram map.
\item \textbf{Analysis.} Every readout lies in an $M$-dimensional reproducing-kernel subspace, and a pointwise bound with a coverage term predicts how far beyond the observed window the memory extrapolates.
\item \textbf{Evidence.} On four task families, the encoder ablation isolates the functional bottleneck as the main source of accuracy, and fixed-checkpoint measurements quantify the retained state.
\end{itemize}

%% file: sections/related_work.tex
\section{Related Work}
\label{sec:related}

\paragraph{Irregular time-series models.}
Latent ODEs, Neural CDEs, and CRU evolve a latent state between observation and query times, and mTAN and ContiFormer relate query times to observations through time-aware attention \citep{rubanova2019latent,kidger2020neural,schirmer2022cru,shukla2021mtan,chen2023contiformer}. Forecasting models reorganize irregular observations through warping, patches, graphs, or points \citep{zhang2023warpformer,yalavarthi2024grafiti,zhang2024tpatchgnn,hu2025vimts,liu2026apn,liu2026astgi}, and TFMixer summarizes them with frequency-domain features \citep{qiu2026tfmixer}. These models answer a requested time by integrating a latent state or attending over observation tokens; HiPPO-FM answers it by evaluating a fixed-rank function.

\paragraph{Decoding at arbitrary inputs.}
Neural processes encode a context set and decode at arbitrary targets: CNP pools the context into one vector, ANP attends from each target to the context, and ConvCNP smooths the context onto a grid with a kernel before decoding \citep{garnelo2018cnp,kim2019anp,gordon2020convcnp}. mTAN attends from a fixed set of reference times to the observations \citep{shukla2021mtan}, and DeepTime fits a closed-form ridge regression on learned time features \citep{woo2023deeptime}. In HiPPO-FM the retained vector has a specified meaning: its entries are interdomain functional values in a HiPPO--Legendre basis, read through the matched inverse-Gram map, so the set of functions the readout can produce is fixed and characterized (Section~\ref{sec:characterization}).

\paragraph{Compact sequence memory.}
Linear attention, latent-bottleneck Transformers, state-space models, and HiPPO compress a sequence into fixed-size states that are updated along the input and read at its positions \citep{katharopoulos2020transformers,jaegle2022perceiverio,feng2024cmanp,smith2023s5,gu2020hippo}; HiPPO-FM reads its state at continuous coordinates.

\paragraph{Interdomain and functional memory.}
Interdomain Gaussian processes define inducing variables as linear functionals of a latent function and predict through their cross-covariances \citep{lazaro2009interdomain}. OHSVGP uses time-varying HiPPO projections as interdomain inducing variables of an online sparse variational GP \citep{NEURIPS2025_c4025018}, Interdomain Attention maintains basis-function coefficients with an SSM recurrence and reads them with query feature maps \citep{kiyohara2026interdomain}, and Functional Attention replaces softmax affinities with linear operators between adaptive bases \citep{xiao2026functional}. HiPPO-FM shares the HiPPO--Legendre functionals and inverse-Gram readout of OHSVGP, stores deterministic neural values trained end to end, and applies self-attention over the observed support before a single write.

%% file: sections/method.tex
\section{Methodology}
\label{sec:method}

\subsection{Problem Setup and Overview}
\paragraph{Irregular observations and queries.}
An irregular observation history is the ordered sequence
\begin{equation}
\label{eq:problem-data}
\mathcal D_P=\{(t_i,x_i,m_i^t)\}_{i=1}^{N},
\qquad
\mathbf t_P=(t_i)_{i=1}^{N},
\qquad
\mathbf r_R=(r_a)_{a=1}^{N_R},
\end{equation}
where $t_i$ is an observed coordinate, $x_i\in\R^{d_x}$ is the zero-filled
measurement after preprocessing, and $m_i^t$ marks a valid position.
Coordinate-valued tasks predict $\widehat Y_R\in\R^{N_R\times d_y}$ on the
requested coordinates. We write $P$ and $R$ for the ordered supports
$\mathbf t_P$ and $\mathbf r_R$; in forecasting, only the permitted history is
encoded.

\paragraph{Shared temporal coordinates.}
Observations and queries are mapped by one affine transform computed from the
valid observed coordinates,
\begin{equation}
\label{eq:coordinate-map}
a_P(t)=\frac{t-t_{\min}}{s_P},
\qquad t_{\min}=\min_{i:m_i^t=1}t_i,
\qquad s_P=\max_{i:m_i^t=1}t_i-t_{\min}.
\end{equation}
The transform depends only on the observed support, so the same requested time
has the same coordinate in every query set; extrapolation queries satisfy
$a_P(r)>1$.

\paragraph{Functional-memory architecture.}
HiPPO-FM separates history encoding from coordinate evaluation:
\begin{equation}
\label{eq:functional-architecture}
S_P=E(\mathcal D_P),
\qquad
\widehat{\mathbf y}(r)=D(S_P,a_P(r))\in\R^{d_y},\qquad r\in R.
\end{equation}
The encoder $E$ applies observed-support self-attention followed by an FM
operator on $P\to P$ (Figure~\ref{fig:hippo_fm}a). For coordinate-valued tasks,
the decoder $D$ is an independent output FM head that reads the encoded history
on $R$; sequence-level tasks pool the encoded observed support instead.

\subsection{Coordinate-Queryable Functional Memory}
\label{sec:fm}
An FM operator takes a representation $H_P\in\R^{N\times d}$ on the support
$P$, writes projected values into a functional memory, and evaluates the
retained coefficients on $R$. It uses $n_h$ heads of width $d_v$, with
$d=n_h d_v$.

\paragraph{Functional coordinate geometry.}
The coordinate system is the shifted Legendre family underlying HiPPO-LegS \citep{gu2020hippo},
\begin{equation}
\label{eq:shifted-legendre}
\phi_m(s)=\sqrt{2m+1}\,L_m(2s-1),
\qquad m=0,\ldots,M-1,
\end{equation}
where $L_m$ is the degree-$m$ Legendre polynomial on $[-1,1]$. On a fixed reference grid $\mathcal S=\{s_g\}_{g=1}^{G}\subset[0,1]$ we set $\Phi_{g,m}=\phi_m(s_g)$, $\Phi\in\R^{G\times M}$. For a positive-semidefinite kernel $k_\theta$, the functional features and Gram matrix are
\begin{equation}
\label{eq:geometry}
\Psi_\theta(X)=\frac{1}{G}K_\theta(X,\mathcal S)\Phi,
\qquad
\psi_\theta(r)=\frac{1}{G}\Phi^\top K_\theta(\mathcal S,r),
\end{equation}
\begin{equation}
\label{eq:functional-gram}
G_\theta=\frac{1}{G^2}\Phi^\top K_\theta(\mathcal S,\mathcal S)\Phi,
\qquad
\bar G_\theta=G_\theta+\epsilon I_M,
\end{equation}
where each row of $\Psi_\theta(X)$ is $\psi_\theta(x)^\top$ and $\epsilon>0$ is numerical jitter. The same geometry is used to write the memory and to read it.

This geometry has a direct interdomain interpretation. For $f\sim\mathcal{GP}(0,k_\theta)$ and the finite-grid functionals $u=|\mathcal S|^{-1}\Phi^\top f(\mathcal S)$, one has $\operatorname{Cov}(f(X),u)=\Psi_\theta(X)$ and $\operatorname{Cov}(u,u)=G_\theta$ \citep{lazaro2009interdomain}. OHSVGP uses HiPPO-derived variables of this kind for online sparse variational inference \citep{NEURIPS2025_c4025018}; here the same identities fix the geometry of a deterministic neural memory.

\paragraph{Functional write.}
Let $D_P=\operatorname{diag}(m_1^t,\ldots,m_N^t)$. For head $h$, an affine
value projection produces $V_{\mathrm{FM},P}^h\in\R^{N\times d_v}$, which is
contracted against the observed features:
\begin{equation}
\label{eq:write}
V_{\mathrm{FM},P}^h=H_PW_{\mathrm{FM},V}^h+\mathbf 1_N(b_{\mathrm{FM},V}^h)^\top,
\qquad
\widetilde C_P^h=\Psi_\theta(P)^\top D_PV_{\mathrm{FM},P}^h.
\end{equation}
The contraction replaces the observation axis of length $N$ with the functional
axis of length $M$. The memory is then normalized and solved into coefficients,
\begin{equation}
\label{eq:memory-solve}
 C_P^h=\operatorname{LN}_{(M,d_v)}(\widetilde C_P^h),
\qquad
\bar G_\theta B_P^h=C_P^h,
\end{equation}
where $\operatorname{LN}_{(M,d_v)}$ is layer normalization \citep{ba2016layer} over the full $M\times d_v$ tensor and the regularized Gram solve removes the correlation between functional coordinates.

\paragraph{Coordinate read.}
Each head evaluates its coefficients at the requested coordinates,
\begin{equation}
\label{eq:read}
Y_R^h=\Psi_\theta(R)B_P^h=\Psi_\theta(R)(G_\theta+\epsilon I_M)^{-1}C_P^h.
\end{equation}
This has the inverse-Gram form of the interdomain sparse-GP predictive mean $K_{fu}K_{uu}^{-1}\mu_u$ \citep{lazaro2009interdomain,titsias2009variational,NEURIPS2025_c4025018}. The role of $\mu_u$ is taken by $C_P^h$, a normalized aggregation of learned values over the observed support; in the sparse GP this slot holds the variational posterior mean. Heads are concatenated and projected,
\begin{equation}
\label{eq:fm-output}
\FM_{P\rightarrow R}(H_P)
=\operatorname{Concat}_{h=1}^{n_h}\!\left[Y_R^h\right]W_O^{\mathrm{FM}}.
\end{equation}

\subsection{Content-Conditioned Encoder and Task Readouts}
\label{sec:encoder}
\paragraph{Content interaction before compression.}
The functional write in Equation~\eqref{eq:write} aggregates values with
weights fixed by the coordinates alone, so any content-dependent interaction
between observations must happen before it. HiPPO-FM therefore applies masked
multi-head self-attention on the observed support
\citep{vaswani2017attention}:
\begin{equation}
\label{eq:mha-qkv}
\begin{aligned}
Q_P^h&=H_PW_Q^h,\quad K_P^h=H_PW_K^h,\quad V_P^h=H_PW_V^h,\\
\boldsymbol{\alpha}_P^h&=\operatorname{softmax}\!\left(
\frac{Q_P^h(K_P^h)^\top}{\sqrt{d_k}}+\mathcal M_P
\right),\qquad
O_P^h=\boldsymbol{\alpha}_P^hV_P^h,\\
U_P&=\operatorname{Concat}_{h=1}^{n_h}\!\left[O_P^h\right]W_O^{\mathrm{MHA}},
\end{aligned}
\end{equation}
where $\mathcal M_P$ masks invalid and padded positions. A forecasting encoder receives only the permitted prefix, so this padding mask
also enforces causality.
Attention decides which observations inform each other, and $\Psi_\theta$
decides where in time the result is stored. This differs from ContiFormer,
which builds continuous-time queries, keys, and values \citep{chen2023contiformer}.

The default MHA-to-FM encoder writes the content-updated representation into
the encoder FM operator,
\begin{equation}
\label{eq:encoder}
H_P^{+}=H_P+\lambda U_P,
\qquad
F_P=\FM^{\mathrm{enc}}_{P\rightarrow P}(H_P^{+}),
\qquad
Z_P=\operatorname{LN}(H_P^{+}+F_P),
\end{equation}
with a learned scalar $\lambda$, and is wrapped in one post-normalization
Transformer block,
\begin{equation}
\label{eq:outer-block}
H_{\mathrm{enc},P}
=\operatorname{LN}\!\left(Z_P+\operatorname{FFN}(Z_P)\right).
\end{equation}
Validity masks apply to attention and to the contraction; the token
normalizations above are separate from the memory normalization in
Equation~\eqref{eq:memory-solve}.

\paragraph{Task-specific readout.}
Coordinate-valued tasks use an independent output FM head with an affine
per-coordinate output layer,
\begin{equation}
\label{eq:task-readout}
\widehat Y_R=\operatorname{Head}^{\mathrm{out}}\!\left(
\FM^{\mathrm{out}}_{P\rightarrow R}(H_{\mathrm{enc},P})
\right),
\end{equation}
so the requested support $R$ enters only through this head. Classification and
next-event prediction ask for a sequence-level output rather than values at
requested times. UEA classification uses last-valid pooling of
$H_{\mathrm{enc},P}$, and next-event prediction applies the intensity head of
the reference setting to the encoded representation at each event
(Appendix~\ref{app:event-protocol}); both read $H_{\mathrm{enc},P}$ directly and bypass $\FM^{\mathrm{out}}$. These
tasks therefore evaluate the content-conditioned encoder, including its
internal FM operator.

\subsection{Functional Semantics and Approximation}
\label{sec:characterization}
Because writing and reading share one kernel-smoothed basis, the write/read
pair defines a finite-dimensional query space. We make this precise and bound
the resulting pointwise error.

\paragraph{Raw-memory semantics.}
For each mode $m$, define the discrete linear functional and its RKHS representer
\begin{equation}
\label{eq:representers}
\mathcal L_m^G f=\frac{1}{G}\sum_{g=1}^{G}\phi_m(s_g)f(s_g),
\qquad
h_m(\cdot)=\frac{1}{G}\sum_{g=1}^{G}\phi_m(s_g)k_\theta(\cdot,s_g).
\end{equation}
Then $h_m(r)=\psi_{\theta,m}(r)$ and $[G_\theta]_{mn}=\langle h_m,h_n\rangle_{\mathcal H_{k_\theta}}$: the coordinate features and the Gram matrix describe the same family of functions.

\begin{proposition}[Raw-memory semantics]
\label{prop:raw-memory-semantics}
For scalar projected values $v\in\R^N$, let $g_{P,v}(\cdot)=\sum_{i=1}^{N}v_i k_\theta(\cdot,t_i)$. Every entry of the unnormalized memory satisfies
\begin{equation}
\label{eq:raw-memory-semantics}
\bigl[\Psi_\theta(P)^\top v\bigr]_m
=\sum_{i=1}^{N}\psi_{\theta,m}(t_i)v_i
=\left\langle g_{P,v},h_m\right\rangle_{\mathcal H_{k_\theta}}
=\mathcal L_m^G g_{P,v}.
\end{equation}
The raw write therefore stores interdomain functional values of the empirical kernel field induced by the observations. Memory normalization changes these values before the Gram solve; the coordinate functions remain fixed.
\end{proposition}

\paragraph{Matched readout and finite query space.}
Write $\Gamma_\epsilon=(G_\theta+\epsilon I_M)^{-1}$ and $\mathcal U_M=\spanof\{h_0,\ldots,h_{M-1}\}$. The inverse is defined for $\epsilon>0$, and for $\epsilon=0$ when $G_\theta$ is nonsingular.

\begin{proposition}[Matched readout and finite query space]
\label{prop:functional}
For fixed kernel, basis, and reference grid, every scalar readout $r\mapsto\psi_\theta(r)^\top \Gamma_\epsilon m$, with $m\in\R^M$, belongs to $\mathcal U_M$. For exact functional values $u_m=\mathcal L_m^G f$, zero jitter, and nonsingular $G_\theta$,
\begin{equation}
\label{eq:matched-projection}
\psi_\theta(r)^\top G_\theta^{-1}u=\bigl(\Pi_{\mathcal U_M}f\bigr)(r),
\qquad f\in\mathcal H_{k_\theta}.
\end{equation}
The Gram inverse maps functional values to the coefficients of their RKHS projection. For an arbitrary learned memory, the coordinate readout remains in a space of dimension at most $M$. Shared geometry across heads and linear channel mixing preserve this span, and affine biases add a constant function, so each scalar output of the affine coordinate head lies in $\mathcal U_M+\operatorname{span}\{1\}$. The observed-support encoder retains a separate nonlinear residual path.
\end{proposition}

\paragraph{Coverage and learned-memory mismatch.}
Define the coverage function
\begin{equation}
\label{eq:coverage-function}
c_\epsilon(r)=k_\theta(r,r)-\psi_\theta(r)^\top \Gamma_\epsilon\psi_\theta(r)\geq0.
\end{equation}

\begin{theorem}[Pointwise functional approximation]
\label{thm:coverage}
Let $f\in\mathcal H_{k_\theta}$, let $u_m=\mathcal L_m^G f$, and let $\widehat u\in\R^M$ be any memory supplied to the decoder. For $\epsilon>0$, or $\epsilon=0$ with nonsingular $G_\theta$,
\begin{equation}
\label{eq:coverage-main}
\left|f(r)-\psi_\theta(r)^\top \Gamma_\epsilon\widehat u\right|
\leq
\norm{f}_{\mathcal H_{k_\theta}}\sqrt{c_\epsilon(r)}
+\norm{\psi_\theta(r)^\top \Gamma_\epsilon}_2\norm{\widehat u-u}_2.
\end{equation}
The first term is the approximation error of the functional geometry at $r$; the second is the mismatch between learned and exact functional values, scaled by the readout norm.
\end{theorem}

The bound separates what the coordinate system can represent from how well the encoder estimates the memory. For a fixed geometry, pre-write content interaction and memory normalization act only on the second term and leave the query space unchanged. The statement concerns one fixed geometry; separately trained models may also differ through their learned kernels and optimization trajectories.

\paragraph{Extrapolation range.}
The reference grid $\mathcal S$ lies in $[0,1]$, so a query with $a_P(r)>1$ reaches the memory only through the kernel tail $k_\theta(r,s_g)$. As $r$ moves away from the grid, $\psi_\theta(r)\to0$ and $c_\epsilon(r)\to k_\theta(r,r)$: the coordinate readout tends to zero and the first term of Equation~\eqref{eq:coverage-main} rises to the prior scale of $f$. The range over which the memory extrapolates is therefore set by the learned kernel lengthscale relative to the observed window. On PhysioNet the hidden window is as long as the observed one, so queries reach $a_P(r)\approx2$ (Section~\ref{sec:physionet}).

\subsection{Functional State and Computational Cost}
\label{sec:reuse}
\begin{table}[t]
\centering
\caption{Cost per history with $N$ observations and $N_R$ requested times, for the models as implemented in our Spiral experiments ($M\le\min(N,d)$, $G=O(d)$, fixed depth and heads). $T_R$ and $T$: adaptive ODE/CDE solver steps up to the requested times; $S$: steps per pairwise ODE in ContiFormer. Seq. steps: solver steps that run in sequence over time. State excludes parameters and workspace. ms/it: measured Spiral training iteration (batch 64, one RTX~5090, five seeds; Table~\ref{tab:wallclock-full}).}
\label{tab:complexity}
\scriptsize
\setlength{\tabcolsep}{2.6pt}
\renewcommand{\arraystretch}{1.05}
\begin{tabularx}{\linewidth}{@{}l|*{4}{>{\centering\arraybackslash}X}|c@{}}
\toprule
Model & Encoding & Query set $R$ & Seq. steps & State & ms/it\\
\midrule
Transformer & $O(Nd)$ & $O(N_R(N_R{+}N{+}d)d)$ & $O(1)$ & $O(Nd)$ & 1.6\\
Neural ODE & $O(Nd^2)$ & $O((T_R{+}N_R)d^2)$ & $O(N{+}T_R)$ & $O(d)$ & 132\\
Latent ODE & $O(Nd^2)$ & $O((T_R{+}N_R)d^2)$ & $O(N{+}T_R)$ & $O(d)$ & 167\\
Neural CDE & $O(Nd)$ & $O(Td^2)$ & $O(T)$ & $O(Nd)$ & 349\\
ContiFormer & -- & $O(N_R^2Sd^2)$ & $O(S)$ & $O(Nd)$ & 3584\\
\midrule
\rowcolor{rowours}\textbf{HiPPO-FM (ours)} & $O(N^2d{+}Nd^2)$ & $O(N_Rd^2)$ & $O(1)$ & $O(Md)$ & 4.6\\
\bottomrule
\end{tabularx}
\end{table}
HiPPO-FM retains $S_P=(B_P,t_{\min},s_P)$ after encoding, of size $O(Md)$. Table~\ref{tab:complexity} lists per-history costs of the Spiral baselines as implemented. The Transformer baseline embeds the observations with a single linear layer and has the cheapest iteration. The ODE and CDE models integrate a latent state with adaptive solvers, so their sequential steps grow with the history and the horizon; ContiFormer interpolates the observations onto the requested times and solves one ODE per token pair for every query set. These solves account for the 29--776$\times$ longer iterations. HiPPO-FM evaluates a new query set in $O(N_Rd^2)$ from $S_P$ alone (Figure~\ref{fig:hippo_fm}d), and its Gram factorization is computed once per model (Appendix~\ref{app:implementation}).

%% file: sections/experiments.tex
\section{Experiments}
\label{sec:experiments}
The experiments ask whether writing the history into a functional coordinate system improves accuracy on tasks that use the memory in different ways: Spiral reconstruction and PhysioNet extrapolation read values at requested times (Sections~\ref{sec:spiral} and~\ref{sec:physionet}), whereas UEA classification and event prediction read the encoded observations (Sections~\ref{sec:uea} and~\ref{sec:event-prediction}). Section~\ref{sec:efficiency} ablates the functional memory and measures cost. Unless stated otherwise, HiPPO-FM uses one encoder block, four heads, width 64 (16 on Spiral), rank $M=16$, shifted-Legendre coordinates, and a learned squared-exponential kernel. Preprocessing, optimization, model selection, and detailed results are in Appendices~\ref{app:experiments}--\ref{app:diagnostics}.

\subsection{Continuous-Coordinate Reconstruction}
\label{sec:spiral}
We first use the two-dimensional Spiral benchmark of ContiFormer
\citep{chen2023contiformer}, comparing HiPPO-FM with ContiFormer, a Transformer,
Neural ODE, Latent ODE, and Neural CDE
\citep{vaswani2017attention,chen2018neural,rubanova2019latent,kidger2020neural}.
Each model observes noisy irregular samples from the first half of a trajectory
and is evaluated by interpolation on that half and extrapolation on the
withheld half, over five seeds (Table~\ref{tab:spiral-main}). All models train for 5,000 iterations, and the checkpoint with the lowest MAE on an independent set of 50 validation trajectories is tested once. Following the ContiFormer Spiral protocol, HiPPO-FM receives the embedded observations linearly interpolated onto the 150-point evaluation grid, and the Transformer, ODE, and CDE baselines receive the 30 raw observations (Appendix~\ref{app:spiral-protocol}).

\begin{table}[!tb]
\centering
\caption{Interpolation and extrapolation on 2-D spirals (mean $\pm$ sample SD over five seeds; lower is better). \textbf{Bold}: best; \underline{underline}: second best. $^\dagger$Archived run of the released implementation, with test-set checkpoint selection. Full settings are in Appendix~\ref{app:spiral-results}.}
\label{tab:spiral-main}
\scriptsize
\setlength{\tabcolsep}{3.2pt}
\renewcommand{\arraystretch}{1.08}
\begin{tabularx}{\linewidth}{@{}l|*{2}{>{\centering\arraybackslash}X}|*{2}{>{\centering\arraybackslash}X}|*{2}{>{\centering\arraybackslash}X}@{}}
\toprule
\multirow{2}{*}{Model} & \multicolumn{2}{c|}{$\rho=0.02$} & \multicolumn{2}{c|}{$\rho=0.2$} & \multicolumn{2}{c}{$\rho=2$}\\
 & RMSE & MAE & RMSE & MAE & RMSE & MAE\\
\midrule
\rowcolor{grouprow}\multicolumn{7}{c}{\texttt{Interpolation} ($\times10^{-2}$)}\\
\midrule
Transformer & 0.95$\smallpm$0.21 & 0.95$\smallpm$0.17 & 2.09$\smallpm$0.72 & 1.74$\smallpm$0.29 & 2.49$\smallpm$0.41 & 2.32$\smallpm$0.25\\
Neural ODE & 1.03$\smallpm$0.19 & 1.01$\smallpm$0.14 & 1.89$\smallpm$0.25 & 1.65$\smallpm$0.10 & 2.18$\smallpm$0.14 & 2.10$\smallpm$0.18\\
Latent ODE & 1.16$\smallpm$0.25 & 1.14$\smallpm$0.20 & 2.50$\smallpm$0.94 & 2.04$\smallpm$0.47 & 2.32$\smallpm$0.39 & 2.14$\smallpm$0.25\\
Neural CDE & 1.32$\smallpm$0.95 & 1.31$\smallpm$0.91 & 2.29$\smallpm$1.59 & 1.89$\smallpm$1.17 & 2.24$\smallpm$0.94 & \underline{1.83$\smallpm$0.40}\\
ContiFormer$^\dagger$ & \underline{0.69$\smallpm$0.16} & \underline{0.70$\smallpm$0.13} & \textbf{0.58$\smallpm$0.14} & \underline{0.55$\smallpm$0.11} & \underline{2.06$\smallpm$2.39} & \underline{1.83$\smallpm$1.98}\\
\rowcolor{rowours}\textbf{HiPPO-FM (ours)} & \textbf{0.42$\smallpm$0.08} & \textbf{0.42$\smallpm$0.09} & \underline{0.61$\smallpm$0.22} & \textbf{0.51$\smallpm$0.11} & \textbf{0.69$\smallpm$0.22} & \textbf{0.60$\smallpm$0.15}\\
\midrule
\rowcolor{grouprow}\multicolumn{7}{c}{\texttt{Extrapolation} ($\times10^{-2}$)}\\
\midrule
Transformer & 1.29$\smallpm$0.11 & 1.29$\smallpm$0.10 & 3.47$\smallpm$1.45 & 3.02$\smallpm$1.10 & 5.23$\smallpm$1.08 & 4.48$\smallpm$0.86\\
Neural ODE & 1.24$\smallpm$0.07 & 1.16$\smallpm$0.09 & 4.13$\smallpm$0.94 & 3.28$\smallpm$0.68 & 4.18$\smallpm$1.03 & 3.61$\smallpm$0.82\\
Latent ODE & 1.18$\smallpm$0.08 & 1.12$\smallpm$0.06 & 4.72$\smallpm$0.43 & 3.68$\smallpm$0.23 & 4.94$\smallpm$1.05 & 3.97$\smallpm$0.74\\
Neural CDE & 2.39$\smallpm$2.73 & 2.32$\smallpm$2.57 & 2.84$\smallpm$2.50 & 2.46$\smallpm$2.40 & 3.70$\smallpm$2.03 & 3.19$\smallpm$1.58\\
ContiFormer$^\dagger$ & \underline{0.88$\smallpm$0.13} & \underline{0.87$\smallpm$0.10} & \underline{0.93$\smallpm$0.35} & \underline{0.90$\smallpm$0.35} & \underline{3.11$\smallpm$2.89} & \underline{2.69$\smallpm$2.39}\\
\rowcolor{rowours}\textbf{HiPPO-FM (ours)} & \textbf{0.62$\smallpm$0.20} & \textbf{0.57$\smallpm$0.18} & \textbf{0.67$\smallpm$0.23} & \textbf{0.61$\smallpm$0.13} & \textbf{0.80$\smallpm$0.17} & \textbf{0.73$\smallpm$0.08}\\
\bottomrule
\end{tabularx}
\end{table}

\FloatBarrier

\begin{figure}[!b]
\centering
\includegraphics[width=0.90\linewidth]{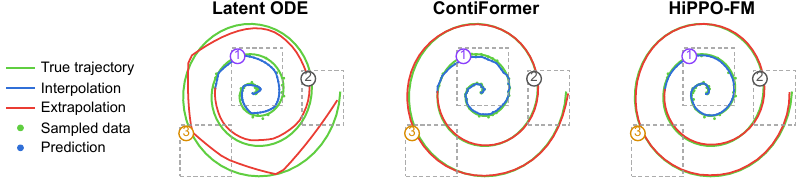}
\caption{Interpolation (first half) and extrapolation (second half) of an irregularly sampled spiral (seed 42, selected checkpoints) by Latent ODE, ContiFormer$^\dagger$, and HiPPO-FM. Dashed boxes 1--3 mark the inner turn, the right endpoint, and the lower-left outer turn.}
\label{fig:spiral-combined}
\end{figure}

HiPPO-FM has the lowest mean error in eleven of the twelve rows of Table~\ref{tab:spiral-main}; ContiFormer has the lowest interpolation RMSE at $\rho=0.2$. The margin grows with noise and with distance from the observations: at $\rho=2$, extrapolation RMSE is $0.80\times10^{-2}$ for HiPPO-FM and $3.11$ for ContiFormer, whose seed SD is $2.89$. In Figure~\ref{fig:spiral-combined}, Latent ODE drifts from the outer turn during extrapolation (box 3), ContiFormer deviates around the inner turn (box 1), and HiPPO-FM follows the true trajectory in all three regions.

\subsection{Irregular Representation Learning}
\label{sec:uea}
We evaluate irregular representation learning on the 20-dataset UEA suite
used by ContiFormer \citep{chen2023contiformer}. The protocol removes
different proportions of observations while retaining their timestamps and
compares recurrent, continuous-time, state-space, and attention-based
baselines under the reproduced data split. The comparison includes GRU-D,
GRU-$\Delta t$ \citep{che2018grud}, ODE-RNN, CADN, Neural CDE, mTAN, S5, TST, and
ContiFormer \citep{che2018grud,rubanova2019latent,chien2023cadn,
kidger2020neural,shukla2021mtan,smith2023s5,zerveas2021tst,
chen2023contiformer}, together with t-PatchGNN
\citep{zhang2024tpatchgnn}.

\begin{table}[!t]
\centering
\caption{Irregular UEA classification over 20 datasets with 30\%, 50\%, and 70\% of observations dropped. ACC is the average accuracy ($\uparrow$) and Rank the average per-dataset rank ($\downarrow$); Overall averages the three settings with equal weight. \textbf{Bold}: best; \underline{underline}: second best. Per-dataset results are in Appendix~\ref{app:uea-full-cells}.}
\label{tab:uea}
\scriptsize
\setlength{\tabcolsep}{4.2pt}
\renewcommand{\arraystretch}{1.05}
\begin{tabularx}{\linewidth}{@{}l|*{2}{>{\centering\arraybackslash}X}|*{2}{>{\centering\arraybackslash}X}|*{2}{>{\centering\arraybackslash}X}|*{2}{>{\centering\arraybackslash}X}@{}}
\toprule
\multirow{2}{*}{Model} & \multicolumn{2}{c|}{30\% dropped} & \multicolumn{2}{c|}{50\% dropped} & \multicolumn{2}{c|}{70\% dropped} & \multicolumn{2}{c}{Overall}\\
 & ACC & Rank & ACC & Rank & ACC & Rank & ACC & Rank\\
\midrule
GRU-D & 0.7140 & 5.50 & 0.7080 & 5.88 & 0.6793 & 6.14 & 0.7004 & 9.00\\
GRU-$\Delta t$ & 0.7255 & 5.35 & 0.7124 & 5.68 & 0.6876 & 5.10 & 0.7085 & 7.00\\
ODE-RNN & 0.7211 & 5.75 & 0.7150 & 5.23 & 0.6949 & 5.20 & 0.7104 & 6.00\\
Neural CDE & 0.5156 & 9.18 & 0.5174 & 9.00 & 0.5028 & 8.90 & 0.5119 & 11.00\\
mTAN & 0.5587 & 8.58 & 0.5447 & 9.23 & 0.5219 & 8.80 & 0.5418 & 10.00\\
CADN & 0.7345 & 5.38 & 0.7251 & 4.85 & 0.6859 & 5.38 & 0.7152 & 4.00\\
S5 & \textbf{0.7497} & \underline{4.70} & \underline{0.7307} & 4.73 & \underline{0.7012} & 5.08 & \underline{0.7272} & \underline{2.00}\\
TST & 0.7378 & 4.78 & 0.7130 & 5.28 & 0.6857 & 5.63 & 0.7122 & 5.00\\
ContiFormer & 0.7257 & 5.18 & 0.7235 & \textbf{4.43} & 0.7010 & \textbf{4.53} & 0.7167 & 3.00\\
t-PatchGNN & 0.7314 & 5.88 & 0.6977 & 6.47 & 0.6756 & 6.36 & 0.7011 & 8.00\\
\midrule
\rowcolor{rowours}\textbf{HiPPO-FM (ours)} & \underline{0.7496} & \textbf{4.60} & \textbf{0.7364} & \underline{4.50} & \textbf{0.7058} & \underline{4.55} & \textbf{0.7306} & \textbf{1.00}\\
\bottomrule
\end{tabularx}
\end{table}

HiPPO-FM achieves the highest overall accuracy and the best overall rank among eleven methods (Table~\ref{tab:uea}). It is strongest under heavier dropping, with the highest average accuracy at 50\% and 70\% dropping, $+0.57$ and $+0.46$ points over S5, and matches S5 within $10^{-4}$ at 30\%. It improves on ContiFormer at every drop ratio, by up to $2.4$ points. Classification pools the encoded observed support (Section~\ref{sec:encoder}), so these results measure the content-conditioned encoder; dataset-level comparisons are in Appendix~\ref{app:uea-stats}.

\subsection{Predicting Irregular Event Sequences}
\label{sec:event-prediction}
Given an ordered history of event times and marks, a model predicts the mark
and the occurrence time of the next event. We use one synthetic and five
real-world datasets (Neonate, Traffic, MIMIC, StackOverflow, and BookOrder),
with four-fold cross-validation for Synthetic, Neonate, and Traffic and five
folds for the others, and report log-likelihood (LL), event-type accuracy, and
next-event-time RMSE. HiPPO-FM uses the conditional-intensity head and multitask objective of the
ContiFormer reference implementation and replaces only the history encoder (Appendix~\ref{app:event-protocol}).

Table~\ref{tab:event-prediction} compares classical Hawkes, recurrent marked
point-process, neural Hawkes, and self-attentive or Transformer Hawkes models
\citep{hawkes1971spectra,du2016rmtpp,mei2017neuralhawkes,zhang2020sahp,
zuo2020thp,ma2022nstka}. HiPPO-FM has the highest event-type accuracy on all
five datasets with more than one event type and the highest LL on four of six
datasets. Its next-event-time RMSE is lowest on Traffic, StackOverflow, and
BookOrder, whereas ContiFormer, ODE-RNN, and NSTKA are more accurate in time on
Synthetic, Neonate, and MIMIC, respectively.

\subsection{Real-World Extrapolation}
\label{sec:physionet}
PhysioNet Challenge 2012 tests the coordinate readout outside the observed window \citep{schirmer2022cru}. Each record has 37 time-varying variables; the first 24 hours are observed, and the output FM head predicts the withheld values up to 48 hours, so queries reach $a_P(r)\approx2$ (Section~\ref{sec:characterization}). Against CRU with its released preprocessing and the same 100-epoch budget over five seeds, HiPPO-FM lowers hidden-region MSE from $0.006465$ to $0.005099$ ($21.1\%$), trains $22.8\times$ faster per epoch, and uses $21\times$ less peak memory (Appendix~\ref{app:cru-results}). CRU keeps its native Gaussian likelihood and HiPPO-FM is trained with MSE, so the comparison concerns point predictions.

\begin{table*}[t]
\centering
\caption{Next-event prediction on irregular event sequences: log-likelihood (LL), event-type accuracy (ACC), and next-event-time RMSE. \textbf{Bold}: best; \underline{underline}: second best. Neonate has a single event type, so ACC is trivially 1 and is not ranked.}
\label{tab:event-prediction}
\scriptsize
\setlength{\tabcolsep}{3.4pt}
\renewcommand{\arraystretch}{1.0}
\resizebox{\textwidth}{!}{%
\begin{tabular}{@{}ll|cccccc@{}}
\toprule
Model & Metric & Synthetic & Neonate & Traffic & MIMIC & StackOverflow & BookOrder\\
\midrule
\multirow{3}{*}{HP} & LL ($\uparrow$) & $-3.084\smallpm0.005$ & $-4.618\smallpm0.005$ & $-1.482\smallpm0.005$ & $-4.618\smallpm0.005$ & $-5.794\smallpm0.005$ & $-1.036\smallpm0.000$\\
 & ACC ($\uparrow$) & $0.756\smallpm0.000$ & -- & $0.570\smallpm0.000$ & $0.795\smallpm0.000$ & $0.441\smallpm0.000$ & $0.604\smallpm0.000$\\
 & RMSE ($\downarrow$) & $0.953\smallpm0.000$ & $10.957\smallpm0.012$ & $0.407\smallpm0.000$ & $1.021\smallpm0.000$ & $1.341\smallpm0.000$ & $3.781\smallpm0.000$\\
\cmidrule(l{2pt}r{2pt}){1-8}
\multirow{3}{*}{RMTPP} & LL ($\uparrow$) & $-1.025\smallpm0.030$ & $-2.817\smallpm0.023$ & $-0.546\smallpm0.012$ & $-1.184\smallpm0.023$ & $-2.374\smallpm0.001$ & $-0.952\smallpm0.007$\\
 & ACC ($\uparrow$) & $0.841\smallpm0.000$ & -- & $0.805\smallpm0.002$ & $0.823\smallpm0.004$ & $0.461\smallpm0.000$ & $0.624\smallpm0.000$\\
 & RMSE ($\downarrow$) & $0.369\smallpm0.014$ & $9.517\smallpm0.023$ & $0.337\smallpm0.001$ & $0.864\smallpm0.017$ & $0.955\smallpm0.000$ & $3.647\smallpm0.003$\\
\cmidrule(l{2pt}r{2pt}){1-8}
\multirow{3}{*}{NeuralHP} & LL ($\uparrow$) & $-1.371\smallpm0.004$ & $-2.795\smallpm0.012$ & $-0.643\smallpm0.004$ & $-1.239\smallpm0.027$ & $-2.608\smallpm0.000$ & $-1.104\smallpm0.005$\\
 & ACC ($\uparrow$) & $0.841\smallpm0.000$ & -- & $0.759\smallpm0.001$ & $0.814\smallpm0.001$ & $0.450\smallpm0.000$ & $0.621\smallpm0.000$\\
 & RMSE ($\downarrow$) & $0.631\smallpm0.002$ & $9.614\smallpm0.013$ & $0.358\smallpm0.001$ & $0.846\smallpm0.007$ & $1.022\smallpm0.000$ & $3.734\smallpm0.003$\\
\cmidrule(l{2pt}r{2pt}){1-8}
\multirow{3}{*}{SAHP} & LL ($\uparrow$) & $-0.619\smallpm0.063$ & $-2.646\smallpm0.057$ & $\mathbf{0.372\smallpm0.022}$ & $\underline{-1.110\smallpm0.030}$ & $-2.404\smallpm0.002$ & $\mathbf{-0.304\smallpm0.002}$\\
 & ACC ($\uparrow$) & $0.841\smallpm0.000$ & -- & $0.780\smallpm0.001$ & $0.830\smallpm0.004$ & $0.455\smallpm0.000$ & $0.622\smallpm0.000$\\
 & RMSE ($\downarrow$) & $0.521\smallpm0.055$ & $\underline{9.403\smallpm0.060}$ & $0.337\smallpm0.000$ & $0.851\smallpm0.006$ & $0.963\smallpm0.000$ & $3.676\smallpm0.011$\\
\cmidrule(l{2pt}r{2pt}){1-8}
\multirow{3}{*}{THP} & LL ($\uparrow$) & $-0.589\smallpm0.017$ & $-2.702\smallpm0.045$ & $-0.569\smallpm0.015$ & $-1.137\smallpm0.038$ & $\underline{-2.354\smallpm0.001}$ & $-1.102\smallpm0.052$\\
 & ACC ($\uparrow$) & $0.841\smallpm0.000$ & -- & $\underline{0.818\smallpm0.001}$ & $\underline{0.834\smallpm0.005}$ & $\underline{0.468\smallpm0.000}$ & $0.622\smallpm0.004$\\
 & RMSE ($\downarrow$) & $0.205\smallpm0.006$ & $9.471\smallpm0.045$ & $\underline{0.332\smallpm0.000}$ & $\underline{0.843\smallpm0.017}$ & $0.951\smallpm0.000$ & $3.688\smallpm0.004$\\
\cmidrule(l{2pt}r{2pt}){1-8}
\multirow{3}{*}{NSTKA} & LL ($\uparrow$) & $-1.001\smallpm0.025$ & $-2.747\smallpm0.061$ & $-0.667\smallpm0.027$ & $-1.188\smallpm0.012$ & $-2.406\smallpm0.003$ & $-0.919\smallpm0.019$\\
 & ACC ($\uparrow$) & $\underline{0.842\smallpm0.000}$ & -- & $0.767\smallpm0.001$ & $0.833\smallpm0.003$ & $0.465\smallpm0.000$ & $0.621\smallpm0.000$\\
 & RMSE ($\downarrow$) & $0.379\smallpm0.015$ & $9.502\smallpm0.049$ & $0.346\smallpm0.001$ & $\mathbf{0.842\smallpm0.008}$ & $0.956\smallpm0.000$ & $3.731\smallpm0.007$\\
\cmidrule(l{2pt}r{2pt}){1-8}
\multirow{3}{*}{GRU-$\Delta t$} & LL ($\uparrow$) & $-0.871\smallpm0.050$ & $-2.736\smallpm0.031$ & $-0.613\smallpm0.062$ & $-1.164\smallpm0.026$ & $-2.389\smallpm0.002$ & $-0.915\smallpm0.006$\\
 & ACC ($\uparrow$) & $0.841\smallpm0.000$ & -- & $0.800\smallpm0.004$ & $0.832\smallpm0.007$ & $0.466\smallpm0.000$ & $\underline{0.627\smallpm0.000}$\\
 & RMSE ($\downarrow$) & $0.249\smallpm0.013$ & $9.421\smallpm0.050$ & $0.335\smallpm0.001$ & $0.850\smallpm0.010$ & $\underline{0.950\smallpm0.000}$ & $3.666\smallpm0.016$\\
\cmidrule(l{2pt}r{2pt}){1-8}
\multirow{3}{*}{ODE-RNN} & LL ($\uparrow$) & $-1.032\smallpm0.102$ & $-2.732\smallpm0.080$ & $-0.491\smallpm0.011$ & $-1.183\smallpm0.028$ & $-2.395\smallpm0.001$ & $-0.988\smallpm0.006$\\
 & ACC ($\uparrow$) & $0.841\smallpm0.000$ & -- & $0.812\smallpm0.000$ & $0.827\smallpm0.006$ & $0.467\smallpm0.000$ & $0.624\smallpm0.000$\\
 & RMSE ($\downarrow$) & $0.342\smallpm0.034$ & $\mathbf{9.289\smallpm0.048}$ & $0.334\smallpm0.000$ & $0.865\smallpm0.021$ & $0.952\smallpm0.000$ & $\underline{3.605\smallpm0.004}$\\
\cmidrule(l{2pt}r{2pt}){1-8}
\multirow{3}{*}{mTAN} & LL ($\uparrow$) & $-0.920\smallpm0.036$ & $-2.722\smallpm0.026$ & $-0.548\smallpm0.023$ & $-1.149\smallpm0.029$ & $-2.391\smallpm0.002$ & $-0.980\smallpm0.004$\\
 & ACC ($\uparrow$) & $\underline{0.842\smallpm0.000}$ & -- & $0.811\smallpm0.002$ & $0.832\smallpm0.009$ & $0.466\smallpm0.000$ & $0.620\smallpm0.000$\\
 & RMSE ($\downarrow$) & $0.286\smallpm0.008$ & $9.363\smallpm0.042$ & $0.334\smallpm0.001$ & $0.848\smallpm0.006$ & $\underline{0.950\smallpm0.000}$ & $3.680\smallpm0.015$\\
\cmidrule(l{2pt}r{2pt}){1-8}
\multirow{3}{*}{ContiFormer} & LL ($\uparrow$) & $\underline{-0.617\smallpm0.031}$ & $\underline{-2.700\smallpm0.031}$ & $-0.628\smallpm0.025$ & $-1.162\smallpm0.028$ & $-2.450\smallpm0.003$ & $\underline{-0.310\smallpm0.014}$\\
 & ACC ($\uparrow$) & $0.841\smallpm0.002$ & -- & $0.799\smallpm0.004$ & $0.820\smallpm0.007$ & $0.458\smallpm0.002$ & $0.613\smallpm0.003$\\
 & RMSE ($\downarrow$) & $\mathbf{0.188\smallpm0.006}$ & $9.472\smallpm0.040$ & $0.350\smallpm0.004$ & $0.972\smallpm0.012$ & $1.010\smallpm0.002$ & $3.750\smallpm0.025$\\
\midrule
\rowcolor{rowours} & LL ($\uparrow$) & $\mathbf{-0.224\smallpm0.068}$ & $\mathbf{-2.621\smallpm0.316}$ & $\underline{0.084\smallpm0.106}$ & $\mathbf{-0.863\smallpm0.264}$ & $\mathbf{-1.165\smallpm0.009}$ & $-0.921\smallpm0.103$\\
\rowcolor{rowours} & ACC ($\uparrow$) & $\mathbf{0.922\smallpm0.008}$ & $1.000\smallpm0.000$ & $\mathbf{0.953\smallpm0.004}$ & $\mathbf{0.873\smallpm0.031}$ & $\mathbf{0.748\smallpm0.003}$ & $\mathbf{0.651\smallpm0.007}$\\
\rowcolor{rowours}\multirow{-3}{*}{\textbf{HiPPO-FM (ours)}} & RMSE ($\downarrow$) & $\underline{0.218\smallpm0.014}$ & $9.521\smallpm4.499$ & $\mathbf{0.249\smallpm0.047}$ & $0.887\smallpm0.128$ & $\mathbf{0.658\smallpm0.004}$ & $\mathbf{3.562\smallpm0.413}$\\
\bottomrule
\end{tabular}}
\end{table*}

\FloatBarrier

\subsection{Functional Bottleneck, Training Cost, and Retained State}
\label{sec:efficiency}
\begin{figure}[t]
\centering
\includegraphics[width=\linewidth]{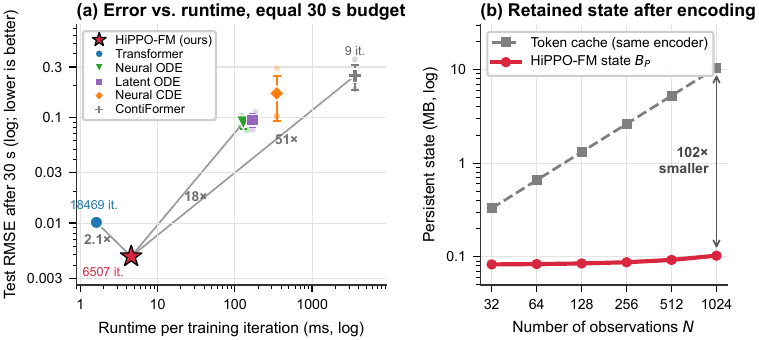}
\vspace{-5mm}
\caption{\textbf{Efficiency diagnostics.} \textbf{(a)} Equal 30\,s training budget on one RTX~5090 (Spiral, five paired seeds): runtime per iteration versus final test RMSE; faint dots are seeds, error bars are sample SD, gray connectors give the RMSE ratio to HiPPO-FM, and labels give completed iterations. \textbf{(b)} Persistent state after encoding for a fixed checkpoint when the same trained encoder is read by the FM head or by a token cross-attention decoder that caches the encoded observations.}
\label{fig:efficiency}
\end{figure}
These experiments use the Spiral protocol of Table~\ref{tab:spiral-main} at $\rho=0.02$ with the same five seeds, paired across variants; the wall-clock runs replace the 5,000-iteration budget with 30\,s of training (Appendix~\ref{app:diagnostics}).

\begin{wraptable}{r}{0.44\linewidth}
\vspace{-4.5mm}
\centering
\caption{Spiral ablation ($\times10^{-2}$, five paired seeds; $k/5$: seeds worse than Full).}
\label{tab:ablation-main}
\scriptsize
\setlength{\tabcolsep}{2.2pt}
\begin{tabular}{@{}lccc@{}}
\toprule
Variant & Overall & Extrap. & $k/5$\\
\midrule
\rowcolor{rowours}Full & $\mathbf{0.54\smallpm0.14}$ & $\mathbf{0.63\smallpm0.20}$ & --\\
w/o encoder FM & $9.88\smallpm0.59$ & $13.79\smallpm0.77$ & 5/5\\
Cosine basis & $0.61\smallpm0.10$ & $0.66\smallpm0.15$ & 4/5\\
No Gram solve & $0.63\smallpm0.07$ & $0.64\smallpm0.07$ & 4/5\\
\bottomrule
\end{tabular}
\vspace{-3mm}
\end{wraptable}
\paragraph{The functional bottleneck.}
Table~\ref{tab:ablation-main} changes one part of the encoder at a time. Removing the encoder FM operator while keeping observed-support self-attention and the feed-forward path raises overall RMSE from $0.54$ to $9.88\times10^{-2}$ and degrades every seed (paired $t$-test $p=3.5\times10^{-6}$; exact Wilcoxon $p=0.0625$, the smallest attainable value for $n=5$ \citep{wilcoxon1945individual}). Variants that keep a rank-$M$ functional memory but change its geometry stay within one seed SD of Full: a cosine basis or omitting the Gram solve raises the mean by less than $0.1\times10^{-2}$ ($p=0.24$ and $0.10$), and point features without the kernel give $0.49\times10^{-2}$ ($p=0.56$; Table~\ref{tab:ablation-full}). The accuracy therefore comes from writing the history into a functional memory; within this family, the HiPPO--Legendre and kernel choices supply the interdomain semantics of Section~\ref{sec:characterization} at no measurable cost in error.

\paragraph{Training cost.}
With shared optimizer settings and a nominal 30\,s budget, HiPPO-FM reaches a test RMSE of $0.483\smallpm0.028\times10^{-2}$, $2.1\times$ lower than the Transformer ($1.015\smallpm0.053\times10^{-2}$) in all five paired seeds (paired $t$-test $p=7.7\times10^{-6}$; exact Wilcoxon $p=0.0625$), although the Transformer completes $2.8\times$ more iterations (Figure~\ref{fig:efficiency}a). Both reach validation MAE $\le0.02$ in every seed, after $3.23\smallpm0.56$\,s and $3.43\smallpm0.30$\,s ($p=0.39$). The ODE, CDE, and ContiFormer baselines need 132--3584\,ms per iteration (Table~\ref{tab:complexity}) and remain above this threshold after 30\,s; ContiFormer completes nine iterations. The comparison is specific to this implementation, budget, and hardware.

\paragraph{Retained state.}
For a batch of 20 histories, the coefficients and coordinate bounds occupy a constant 82.1\,kB; with the $N$-length validity mask (one byte per observation) the measured state grows from 82.7 to 102.6\,kB as $N$ goes from 32 to 1024. The token cache of the same trained encoder grows linearly to 10.5\,MB, $102\times$ larger (Figure~\ref{fig:efficiency}b). The token decoder is faster per call (6.1 vs.\ 8.9\,ms encoding and 0.85 vs.\ 1.63\,ms readout at $N=1024$), and peak transient allocation is similar (719 vs.\ 713\,MB); the fixed rank reduces what must be stored per history.

%% file: sections/conclusion.tex
\section{Conclusion}
HiPPO-FM stores an irregular history as a rank-$M$ function with explicit time coordinates: masked self-attention mixes the observation contents, and the updated values are written through kernel-smoothed HiPPO--Legendre functionals and read through the matched inverse-Gram map. This write/read pair fixes an $M$-dimensional reproducing-kernel readout space, and its coverage term bounds how far the memory can extrapolate. Empirically, the functional memory is the component the model depends on most: removing it from the encoder raises Spiral RMSE 18-fold, and with it HiPPO-FM leads on Spiral reconstruction, irregular UEA classification, multi-type event prediction, and PhysioNet extrapolation. The fixed rank also keeps the retained state at a constant size, $102\times$ smaller than a token cache at 1,024 observations.

%% file: sections/statements.tex
\section*{AI use statement}
In this work, we used generative AI tools, including ChatGPT and Codex, as
auxiliary tools for literature organization, code debugging, mathematical
expression refinement, manuscript language improvement, and proofreading. AI
tools were not used to generate the core research ideas, design the proposed
methodology, formulate the experimental protocols, create or modify
experimental data, or determine scientific conclusions.

All methodological decisions, theoretical analyses, experimental
implementations, result interpretations, and final manuscript contents were
developed and verified by the authors. The authors carefully reviewed all
AI-assisted outputs, including citations, code, figures, and text, and take
full responsibility for the accuracy and integrity of the submitted work.

\section*{Reproducibility statement}
Appendix~\ref{app:implementation} specifies the architecture, hyperparameters, and the encode/query procedure (Algorithm~\ref{alg:hippo-fm}); Appendix~\ref{app:theory} gives the proofs; Appendix~\ref{app:experiments} records the preprocessing, training, selection, and metric protocol of every experiment; Appendix~\ref{app:additional} contains detailed results; and Appendix~\ref{app:diagnostics} reports the ablation and efficiency diagnostics.

\section*{Ethics statement}
The clinical experiments concern retrospective benchmark prediction, not clinical decision-making or deployment. Lower reconstruction error does not establish medical utility, calibration, fairness across patient groups, or safety. The use of evaluation-split statistics in the inherited CRU preprocessing is disclosed and limits deployment-oriented interpretation. Any subsequent clinical use would require independent validation, appropriate data governance, and domain-specific review.

%% file: appendix/method_details.tex
\section{Implementation Details and Algorithm}
\label{app:implementation}

\subsection{Algorithm}
Algorithm~\ref{alg:hippo-fm} summarizes one forward pass and follows
Figure~\ref{fig:hippo_fm}; the hyperparameters are listed in
Appendix~\ref{app:hyperparameters}. The Gram factorization depends only on the
parameters and is computed once per model. \textsc{Encode} is run once per
history; \textsc{Query} reads only the retained state $S_P$, so each additional
query set costs one feature evaluation and one matrix product per head.

\begin{algorithm}[H]
\caption{HiPPO Functional Memory: encode once, query many}
\label{alg:hippo-fm}
\small
\begin{algorithmic}[1]
\Require history $\mathcal D_P=\{(t_i,x_i,m_i^t)\}_{i=1}^{N}$; query sets $R_1,\ldots,R_J$; basis $\Phi\in\R^{G\times M}$ on grid $\mathcal S\subset[0,1]$; kernel $k_\theta$; jitter $\epsilon$
\Ensure predictions $\widehat Y_{R_1},\ldots,\widehat Y_{R_J}$ (or a sequence-level output)
\Statex \textit{Once per model}
\State $G_\theta\gets G^{-2}\,\Phi^\top K_\theta(\mathcal S,\mathcal S)\,\Phi$;\quad $LL^\top\gets\operatorname{chol}(G_\theta+\epsilon I_M)$ \Comment{Eq.~\eqref{eq:functional-gram}}
\Statex
\Procedure{FMWrite}{$H$, $\mathbf a$, $D$} \Comment{$H\in\R^{N\times d}$, coordinates $\mathbf a$, validity $D$}
  \For{each head $h=1,\ldots,n_h$}
    \State $V^h\gets HW_V^h+\mathbf 1_N(b_V^h)^\top$
    \State $\widetilde C^h\gets\Psi_\theta(\mathbf a)^\top D\,V^h$,\quad $\Psi_\theta(\mathbf a)=G^{-1}K_\theta(\mathbf a,\mathcal S)\Phi$ \Comment{Eqs.~\eqref{eq:geometry},~\eqref{eq:write}}
    \State $B^h\gets L^{-\top}L^{-1}\,\operatorname{LN}_{(M,d_v)}(\widetilde C^h)$ \Comment{Eq.~\eqref{eq:memory-solve}}
  \EndFor
  \State \Return $\{B^h\}_{h=1}^{n_h}$ \Comment{$M\times d_v$ per head}
\EndProcedure
\Procedure{FMRead}{$\{B^h\}$, $\mathbf r$}
  \State \Return $\operatorname{Concat}_h\!\big[\Psi_\theta(\mathbf r)B^h\big]\,W_O$ \Comment{Eqs.~\eqref{eq:read},~\eqref{eq:fm-output}}
\EndProcedure
\Statex
\Procedure{Encode}{$\mathcal D_P$}
  \State $t_{\min},s_P\gets$ minimum and range of the valid $t_i$;\quad $a_i\gets(t_i-t_{\min})/s_P$ \Comment{Eq.~\eqref{eq:coordinate-map}}
  \State $H_P\gets\operatorname{InputProj}(x)+\operatorname{TimeEnc}(\mathbf a)$
  \State $H_P^+\gets H_P+\lambda\,\operatorname{MHA}(H_P;\mathcal M_P)$ \Comment{content interaction before the write}
  \State $F_P\gets\textsc{FMRead}(\textsc{FMWrite}(H_P^+,\mathbf a,D_P),\mathbf a)$ \Comment{encoder FM, $P\to P$}
  \State $Z_P\gets\operatorname{LN}(H_P^++F_P)$;\quad $H_{\mathrm{enc},P}\gets\operatorname{LN}(Z_P+\operatorname{FFN}(Z_P))$ \Comment{Eqs.~\eqref{eq:encoder},~\eqref{eq:outer-block}}
  \If{the task is sequence-level}
    \State \Return $\operatorname{Pool}(H_{\mathrm{enc},P})$ \Comment{UEA; event head on each $H_{\mathrm{enc},j}$}
  \EndIf
  \State $B_P\gets\textsc{FMWrite}_{\mathrm{out}}(H_{\mathrm{enc},P},\mathbf a,D_P)$
  \State \Return $S_P=(B_P,t_{\min},s_P)$ \Comment{$O(Md)$; observation tokens are released}
\EndProcedure
\Procedure{Query}{$S_P$, $R$}
  \State $\mathbf r\gets(R-t_{\min})/s_P$ \Comment{entries may exceed 1}
  \State \Return $\operatorname{Head}_{\mathrm{out}}\big(\textsc{FMRead}_{\mathrm{out}}(B_P,\mathbf r)\big)$ \Comment{Eq.~\eqref{eq:task-readout}}
\EndProcedure
\Statex
\State $S_P\gets\textsc{Encode}(\mathcal D_P)$
\For{$j=1,\ldots,J$}
  \State $\widehat Y_{R_j}\gets\textsc{Query}(S_P,R_j)$
\EndFor
\end{algorithmic}
\end{algorithm}

\subsection{Architecture and hyperparameters}
\label{app:hyperparameters}
The main configuration uses hidden width $d=64$, one encoder block, four
attention heads, a two-layer FFN, a shifted-Legendre basis on a reference grid
of size $G=128$, and memory rank $M=16$; the Spiral model uses width 16 with the
same rank. The encoder applies masked self-attention on the observed support
before the encoder FM operator, and the output FM head uses the same functional
construction with separate parameters. Missing channels are zero-filled, and
validity masks are applied to attention and to the functional contraction. The
memory LayerNorm acts over the full $(M,d_v)$ tensor, and classification uses
last-valid pooling. All temporal coordinates use the affine map of
Equation~\eqref{eq:coordinate-map}, computed from the valid observed
coordinates and reused for every query set, so extrapolation queries satisfy
$a_P(r)>1$.

\input{appendix/query_reuse}

%% file: appendix/query_reuse.tex
\subsection{Query reuse and consistency}
\label{app:query-reuse}
\begin{wrapfigure}[11]{r}{0.42\linewidth}
\vspace{-6mm}
\centering
\includegraphics[width=0.98\linewidth]{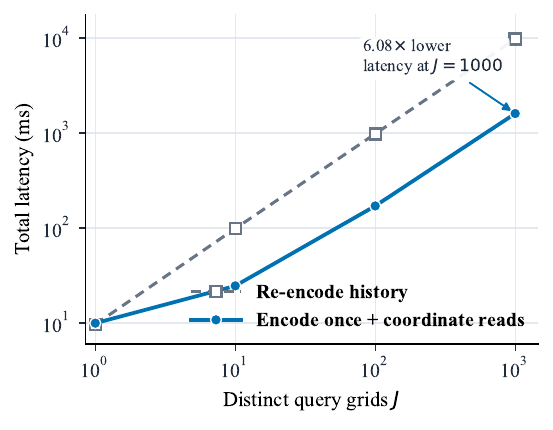}
\caption{Repeated-query latency.}
\label{fig:query-reuse}
\vspace{-3mm}
\end{wrapfigure}
For a fixed observation history, we compare direct evaluation with cached-state
evaluation of Algorithm~\ref{alg:hippo-fm} under different query orderings,
groupings, and chunk sizes. The maximum coordinate-wise difference is
$2.611\times10^{-5}$, so one encoded state can be reused across query sets;
the check covers a fixed history and fixed parameters. The latency sweep in
Figure~\ref{fig:query-reuse} varies the number $J$ of distinct query grids for
one history. At $J=1000$, encoding once and reusing the retained state reduces
total latency from $9756.26$\,ms to $1605.35$\,ms, a $6.08\times$ reduction.
This is an amortized schedule comparison; the coordinate-read cost still grows
with the number of requested grids.

%% file: appendix/theory.tex
\section{Functional Geometry and Proofs}
\label{app:theory}
We analyze the implemented finite-grid operator; its continuous-functional analogue follows by replacing the reference-grid sums with integrals. For notational compactness, we write $k=k_\theta$, $\mathcal H_k=\mathcal H_{k_\theta}$, and $\psi=\psi_\theta$ unless stated otherwise.

\subsection{Finite-grid representers and covariance consistency}
The finite-grid coordinates are bounded RKHS linear functionals, following the interdomain-GP construction \citep{lazaro2009interdomain}. OHSVGP applies this perspective to time-varying HiPPO projections in online sparse variational inference \citep{NEURIPS2025_c4025018}. The following derivation establishes the RKHS representers and Gram identities used by the deterministic HiPPO-FM operator.

Define bounded linear functionals
\begin{equation}
\label{eq:app-representers}
\mathcal L_m^G f=G^{-1}\sum_{g=1}^G\phi_m(s_g)f(s_g),\qquad
h_m=G^{-1}\sum_{g=1}^G\phi_m(s_g)k(\cdot,s_g).
\end{equation}
Point evaluation is bounded in an RKHS; finite linear combinations are therefore bounded. The reproducing property gives $\mathcal L_m^G f=\langle f,h_m\rangle_{\mathcal H_k}$. Define the synthesis operator $\mathsf H:\R^M\to\mathcal H_k$ by $\mathsf H a=\sum_m a_m h_m$. Then
\begin{equation}
G_\theta=\mathsf H^*\mathsf H=G^{-2}\Phi^\top K(\mathcal S,\mathcal S)\Phi,
\qquad
\psi(r)=\mathsf H^*k(r,\cdot)=G^{-1}\Phi^\top K(\mathcal S,r).
\label{eq:app-gram}
\end{equation}
These identities show that the write features and Gram matrix use the same RKHS representers. Thus $G_\theta=\mathsf H^\ast\mathsf H\succeq0$, and positive jitter yields a positive-definite coefficient solve.

For a conceptual zero-mean GP prior $f\sim\mathcal{GP}(0,k)$, covariance bilinearity gives $\operatorname{Cov}(f(r),u)=\psi(r)^\top$ and $\operatorname{Cov}(u,u)=G_\theta$ for the functionals in Equation (20) \citep{lazaro2009interdomain,NEURIPS2025_c4025018}.
\begin{equation}
\operatorname{Cov}(f(r),u)=\psi(r)^\top,
\qquad \operatorname{Cov}(u,u)=G_\theta.
\end{equation}
With invertible $G_\theta$, exact functional observations give the standard GP conditional mean $\psi(r)^\top G_\theta^{-1}u$, while sparse variational prediction has the form $m_q(r)=K_{fu}(r)K_{uu}^{-1}\mu_u$ \citep{titsias2009variational}. In HiPPO-FM, these identities characterize the coordinate geometry; the learned state is formed from projected neural values and decoded through the regularized solve.

\subsection{Raw-memory semantics}
\paragraph{Proof of Proposition~\ref{prop:raw-memory-semantics}.}
Let $v\in\R^N$ and $g_{P,v}=\sum_i v_i k(\cdot,t_i)$. For every functional coordinate, the reproducing property gives
\begin{equation}
[\Psi(P)^\top v]_m=\sum_i v_i h_m(t_i)
=\left\langle\sum_i v_i k(\cdot,t_i),h_m\right\rangle_{\mathcal H_k}
=\mathcal L_m^Gg_{P,v}.
\end{equation}
This is the functional value of the empirical kernel field induced by $v$.

\subsection{Matched readout and finite query space}
Write $\Gamma_\epsilon=(G_\theta+\epsilon I_M)^{-1}$ and $\mathcal U_M=\spanof\{h_0,\ldots,h_{M-1}\}$. The inverse is defined for $\epsilon>0$, and also for $\epsilon=0$ when $G_\theta$ is nonsingular.

\paragraph{Proof of Proposition~\ref{prop:functional}: matched projection.}
Assume $G_\theta$ is nonsingular, let $u=\mathsf H^*f$, and set $s=\mathsf H G_\theta^{-1}u$. For each $n$,
\begin{equation}
\langle f-s,h_n\rangle=u_n-[G_\theta G_\theta^{-1}u]_n=0.
\end{equation}
Since $s\in\mathcal U_M$ and $f-s$ is orthogonal to every element of $\mathcal U_M$, $s=\Pi_{\mathcal U_M}f$. Evaluation gives $s(r)=\psi(r)^\top G_\theta^{-1}u$. The inverse corrects the Gram geometry of the representers; replacing it with the identity is equivalent only in the special case where those representers are orthonormal.

\paragraph{Arbitrary normalized memory.}
For any $m$, let $\Gamma_\epsilon=(G_\theta+\epsilon I)^{-1}$. Then
\begin{equation}
o_m(r)=\psi(r)^\top \Gamma_\epsilon m
=\sum_j [\Gamma_\epsilon m]_j h_j(r)\in\mathcal U_M.
\end{equation}
LayerNorm changes $m$, not the functions $h_j$. With shared geometry, concatenation and linear channel mixing preserve this functional span. An affine output bias adds a constant function, yielding the upper bound $\mathcal U_M+\spanof\{1\}$. A nonlinear task head would not generally preserve this span; the natural coordinate-prediction head considered here is affine. The observed-support encoder has its own nonlinear residual path and is not itself asserted to have rank $M$.

\subsection{Induced raw kernel and finite-rank compression}
\begin{proposition}[Raw induced kernel]
The raw, unnormalized coordinate operator has kernel
\begin{equation}
\label{eq:raw-coordinate-attention}
k_{M,\epsilon}(r,t)=\psi(r)^\top \Gamma_\epsilon\psi(t),
\qquad o_{\rm raw}(r)=\sum_i k_{M,\epsilon}(r,t_i)v_i.
\end{equation}
It is positive semidefinite and has rank at most $M$ on any finite coordinate set.
\end{proposition}
\begin{proof}
For a matrix $\Psi_R$ of query features, $K_M(R,R)=\Psi_R \Gamma_\epsilon\Psi_R^\top$. For any vector $z$, $z^\top K_Mz=\norm{\Gamma_\epsilon^{1/2}\Psi_R^\top z}_2^2\ge0$. The product factors through an $M$-dimensional space, so its rank is at most $M$. Substitution of the raw write gives the displayed prediction.
\end{proof}

\begin{proposition}[Collisions of the raw writer]
If $\delta\in\R^{N\times d_v}$ satisfies $\Psi(P)^\top\delta=0$, then $V$ and $V+\delta$ have identical raw memories. Applying the same deterministic memory normalization and decoder preserves this equality in exact arithmetic.
\end{proposition}
\begin{proof}
$\Psi(P)^\top(V+\delta)=\Psi(P)^\top V$. Equal inputs remain equal under deterministic functions.
\end{proof}
The condition applies at the input of the \emph{linear contraction}, not automatically to raw sensor inputs before arbitrary nonlinear encoders. For a pre-write transform $F$, the relevant difference becomes $\Psi(P)^\top[F(H+\delta)-F(H)]$. It need not vanish, but it also need not be nonzero. For example, the identity retains the collision; a nonlinear MLP or MHA can change it. None of these statements orders their downstream risks or proves injectivity of the compressed representation.

\subsection{Proof of Theorem~\ref{thm:coverage}}
\begin{proof}
Write the error as $[f(r)-\psi(r)^\top \Gamma_\epsilon u]+\psi(r)^\top \Gamma_\epsilon(u-\widehat u)$. The second term is bounded by Euclidean Cauchy--Schwarz. Define $q_r=k(r,\cdot)-\mathsf H \Gamma_\epsilon\psi(r)$. The first term is $\langle f,q_r\rangle_{\mathcal H_k}$, and
\begin{align}
\norm{q_r}_{\mathcal H_k}^2
&=k(r,r)-2\psi(r)^\top \Gamma_\epsilon\psi(r)
+\psi(r)^\top \Gamma_\epsilon G_\theta \Gamma_\epsilon\psi(r)\\
&=c_\epsilon(r)-\epsilon\psi(r)^\top \Gamma_\epsilon^2\psi(r)
\le c_\epsilon(r),
\end{align}
using $\Gamma_\epsilon G_\theta \Gamma_\epsilon=\Gamma_\epsilon-\epsilon \Gamma_\epsilon^2$. This also proves $c_\epsilon(r)\ge0$. RKHS Cauchy--Schwarz completes the bound.
\end{proof}
The first term measures functional geometry; the second measures coefficient mismatch amplified by the readout. The network's normalized memory is one possible $\widehat u$. The theorem bounds representation error for fixed geometry; the effects of attention and learned rank on optimization and test error lie outside its scope. When $\widehat u=\mathsf H^*g_{P,v}$, it describes approximation of the empirical field; that field need not equal the target trajectory.

For a perturbed decoder $\widehat D_r$ in place of $D_r=\psi(r)^\top \Gamma_\epsilon$, triangle inequality adds $\norm{\widehat D_r-D_r}_2\norm{\widehat u}_2$ to Equation~\eqref{eq:coverage-main}. This separates a numerical decoder discrepancy from representational error without assuming a particular floating-point implementation.

\subsection{Query consistency and reuse}
\begin{proposition}[Query-set consistency]
\label{prop:query}
Fix the encoded state $S_P$. In evaluation mode, the prediction at coordinate $r$ is unchanged by the order, grouping, or chunking of other query coordinates, provided every query uses the same observed-support coordinate map and the post-state decoder acts pointwise in $r$.
\end{proposition}

For the frozen state $B_P$, a row of the output has the form
\begin{equation}
\widehat{\mathbf y}(r)=W_y\left(\operatorname{Concat}_h[\psi(a_P(r))^\top B_P^h]W_O+b_O\right)+b_y.
\end{equation}
No quantity on the right depends on any other query in $R$. Permutation, grouping, and chunking therefore preserve the row corresponding to $r$. This proves Proposition~\ref{prop:query}. The assumptions exclude batch statistics across queries, query self-attention, query-dependent coordinate normalization, and stochastic dropout at evaluation.

Caching $B_P$, $t_{\min}$, $s_P$, and validity changes only the execution schedule. The kernel parameters, reference grid, normalization parameters, and output projections are fixed model parameters at inference. Exact equivalence therefore holds algebraically. Changing any such parameter invalidates a cache, as does changing $P$ or its observed values. The numerical tests evaluate these statements for the implemented path; they do not establish invariance under changes to the observation set.

%% file: appendix/protocols.tex
\section{Experimental Protocols}
\label{app:experiments}
This appendix records the data, training, selection, and evaluation protocol
of each main-text experiment, in the order of Section~\ref{sec:experiments}.
Unless a task-specific exception is stated, all methods share the same
preprocessing and evaluation protocol.

\subsection{Spiral reconstruction}
\label{app:spiral-protocol}
Each seed generates 300 trajectories with 150 equally spaced time positions,
of which 200 are used for training and 100 for testing; an independent set of
50 validation trajectories is generated with seed $s+10^6$. Each training step
samples 30 observation times from the first half and a batch of 64
trajectories; the observation noise scale is 0.1 and $\rho\in\{0.02,0.2,2\}$.
All models use Adam \citep{kingma2015adam} with learning rate $10^{-2}$,
dropout 0, and 5,000 iterations, with the squared error over the 150 grid
positions as training loss. Validation MAE is computed after every iteration,
and the checkpoint with the lowest validation MAE is evaluated once on the test
set. Five seeds $\{13,27,42,100,2024\}$ are used. Following the ContiFormer
Spiral protocol, HiPPO-FM linearly interpolates the embedded observations onto
the 150-point evaluation grid before its encoder block and reads each grid
position; the Transformer, Neural ODE, Latent ODE, and Neural CDE baselines
receive the 30 raw observations and their released readouts. The ContiFormer
row of Table~\ref{tab:spiral-main} is an archived run of the released
implementation, which supervises the full grid and selects checkpoints on the
test set; this selection can only favor ContiFormer. The complete $\rho$ sweep and qualitative
visualization are reported in Appendix~\ref{app:spiral-results}.

\subsection{UEA irregular time-series classification}
\input{appendix/uea_setup}

\subsection{Irregular event-sequence prediction}
\label{app:event-protocol}
Each example is an ordered marked sequence. The model receives the history up
to the current event and predicts the mark and time of the next event. We use
one synthetic and five real-world datasets: Neonate, Traffic, MIMIC,
StackOverflow, and BookOrder. Synthetic, Neonate, and Traffic use four-fold
cross-validation; MIMIC, StackOverflow, and BookOrder use five folds. The
reported metrics are next-event log-likelihood, event-type accuracy, and
next-event-time RMSE. Dataset sizes, event cardinalities, and split details
are given in Appendix~\ref{app:event-details}.

\paragraph{Event model and objective.}
HiPPO-FM replaces only the history encoder of the ContiFormer reference
implementation; the event embedding, prediction head, likelihood, and loss are
those of that implementation. Each event $(t_j,k_j)$ is embedded by its mark,
and the encoder runs on the observed support $\{t_j\}$. The encoded sequence
passes through the reference single-layer LSTM \citep{hochreiter1997long}, giving a history vector $h_j$
after every event. The conditional intensity of mark $k$ after event $j$ has
the Transformer Hawkes form \citep{zuo2020thp},
\begin{equation*}
\lambda_k(t\mid\mathcal H_{t_j})=\operatorname{softplus}_{\beta_k}\!\bigl(w_k^\top h_j+b_k+\alpha_k(t-t_j)\bigr),
\qquad t\in(t_j,t_j+t_{\max}],
\end{equation*}
with $\operatorname{softplus}_\beta(x)=\beta^{-1}\log(1+e^{\beta x})$ and
learned $w_k,b_k,\alpha_k,\beta_k$. With $\lambda=\sum_k\lambda_k$, the
sequence log-likelihood is
\begin{equation*}
\log\mathcal L=\sum_{j}\log\lambda_{k_{j+1}}(t_{j+1}\mid\mathcal H_{t_j})
-\sum_j\int_{t_j}^{t_{j+1}}\lambda(s\mid\mathcal H_{t_j})\,ds,
\end{equation*}
where the integral is approximated on a uniform grid of step $0.01$. The next
time and mark are read from the induced density
$p_j(\tau)=\lambda(t_j+\tau)\exp\!\bigl(-\int_0^\tau\lambda(t_j+s)\,ds\bigr)$:
$\hat\tau_{j+1}=\int_0^{t_{\max}}\tau\,p_j(\tau)\,d\tau$ and
$\hat k_{j+1}=\arg\max_k\int_0^{t_{\max}}\frac{\lambda_k}{\lambda}(t_j+\tau)\,p_j(\tau)\,d\tau$,
both by the trapezoidal rule. Training minimizes
\begin{equation*}
-\log\mathcal L+\gamma_{\mathrm{mark}}\,\mathrm{CE}_{0.1}(\hat k,k)
+\gamma_{\mathrm{time}}^{-1}\sum_j(\hat\tau_{j+1}-\tau_{j+1})^2,
\end{equation*}
where $\mathrm{CE}_{0.1}$ is cross-entropy with label smoothing $0.1$ \citep{szegedy2016rethinking} and
the reference configuration uses $t_{\max}=5$, $\gamma_{\mathrm{mark}}=1$, and
$\gamma_{\mathrm{time}}=100$. LL in Table~\ref{tab:event-prediction} is this
log-likelihood, accuracy compares $\hat k_{j+1}$ with $k_{j+1}$, and RMSE
compares $\hat\tau_{j+1}$ with the observed inter-event time.

\subsection{PhysioNet extrapolation}
\label{app:physionet-protocol}
The comparison with CRU \citep{schirmer2022cru} uses
sets A/B with random-state-0 splits of 4,800/1,600/1,600 records, removes
static attributes, quantizes time to 0.1 hours, normalizes each split
separately as in the released CRU preprocessing, and resets missing values to
zero. Observations at or after 24 hours are hidden and targets are evaluated
over the withheld part of the first 48 hours. Both methods use Adam, batch size
100, a 100-epoch budget, and seeds $1,\ldots,5$; CRU keeps its masked Gaussian
NLL, and HiPPO-FM uses the output FM head with an MSE objective.

\subsection{Metrics}
Spiral reconstruction uses RMSE and MAE on clean target trajectories. UEA
classification reports accuracy and per-dataset rank, with the aggregate
quantities formed from the three observation-drop settings. Event prediction
reports LL, event-type accuracy, and next-event-time RMSE. PhysioNet
extrapolation reports hidden-region MSE. Coordinate normalization is always
computed from the observed support.

%% file: appendix/uea_setup.tex
We use the official TRAIN/TEST partitions for the 20 UEA datasets in
Table~\ref{tab:uea-datasets}. For each dataset, 30\%, 50\%, or 70\% of the
observations are removed at random while retained values remain paired with
their original timestamps. All reproduced methods use seeds
$\{27,42,1024\}$ and a fixed early-stopping protocol: the complete
official TRAIN split is used for optimization, training lasts at most 1,000
epochs, and it stops when training accuracy has not improved for 100
consecutive epochs. Classification is performed directly from the
observed-support representation and does not use the final coordinate
readout. The same training, masking, and evaluation procedure is
used for all three observation-drop settings. The TEST split is reserved for
final reporting and is not used for checkpoint selection.

Results for the 50\% MHA-to-FM setting cover all 20 datasets and three
seeds. The baseline columns use the corresponding full 20-dataset results
under the same early-stopping procedure. Cells marked $\dagger$ retain the
reported exception: TST on LSST uses its reporting substitution, and Neural
CDE on HandMovementDirection at 30\% and 50\% uses the reported finite-
checkpoint accuracy despite a non-finite training loss. No such
exception is used for HiPPO-FM.

\subsubsection{Dataset information}
\label{app:uea-datasets}
\begin{table}[h]
\centering
\scriptsize
\caption{Dataset information for the 20-dataset UEA classification suite.}
\label{tab:uea-datasets}
\setlength{\tabcolsep}{4.2pt}
\begin{tabular*}{\linewidth}{@{\extracolsep{\fill}}lcrrrrr@{}}
\toprule
Dataset & Code & \# Train & \# Test & Length & \# Classes & \# Variables\\
\midrule
ArticularyWordRecognition & AWR & 275 & 300 & 144 & 25 & 9\\
BasicMotions & BM & 40 & 40 & 100 & 4 & 6\\
CharacterTrajectories & CT & 1422 & 1436 & 182 & 20 & 3\\
DuckDuckGeese & DDG & 60 & 40 & 270 & 5 & 1345\\
Epilepsy & EP & 137 & 138 & 207 & 4 & 3\\
ERing & ER & 30 & 270 & 65 & 6 & 4\\
FingerMovements & FM & 316 & 100 & 50 & 2 & 28\\
HandMovementDirection & HMD & 160 & 74 & 400 & 4 & 10\\
Handwriting & HW & 150 & 850 & 152 & 26 & 3\\
Heartbeat & HB & 204 & 205 & 405 & 2 & 61\\
JapaneseVowels & JV & 270 & 370 & 29 & 9 & 12\\
Libras & LB & 180 & 180 & 45 & 15 & 2\\
LSST & LSST & 2459 & 2466 & 36 & 14 & 6\\
NATOPS & NATOPS & 180 & 180 & 51 & 6 & 24\\
PEMS-SF & PEMS & 267 & 173 & 144 & 7 & 963\\
PenDigits & PD & 7494 & 3498 & 8 & 10 & 2\\
RacketSports & RS & 151 & 152 & 30 & 4 & 6\\
SelfRegulationSCP1 & SR & 268 & 293 & 896 & 2 & 6\\
SpokenArabicDigits & SA & 6599 & 2199 & 93 & 10 & 13\\
UWaveGestureLibrary & UGL & 2238 & 2241 & 315 & 8 & 3\\
\bottomrule
\end{tabular*}
\end{table}

%% file: appendix/additional.tex
\section{Additional Experimental Results}
\label{app:additional}
This appendix reports detailed results for the main-text experiments in the
order of Section~\ref{sec:experiments}; the corresponding protocols are in
Appendix~\ref{app:experiments}, and the ablation and efficiency diagnostics
are in Appendix~\ref{app:diagnostics}.

\subsection{Full Spiral settings and visualization}
\label{app:spiral-results}
\input{appendix/historical_spiral}

\subsection{Full UEA results}
\label{app:uea-full-cells}
\input{appendix/uea_details}

\subsection{Dataset-level tests on UEA}
\label{app:uea-stats}
Table~\ref{tab:uea-stats} reports two-sided Wilcoxon signed-rank tests
\citep{wilcoxon1945individual} on the reported three-seed per-dataset means,
with Holm correction over the six comparisons \citep{holm1979simple}. All six
Holm-corrected $p$-values remain above $0.05$; the test therefore describes
cross-dataset ordering rather than paired example-level predictions.

\begin{table}[h]
\centering
\footnotesize
\setlength{\tabcolsep}{5pt}
\caption{Dataset-level comparison of HiPPO-FM with S5 and ContiFormer on UEA.
$\Delta$ACC is the mean accuracy difference in percentage points; W/T/L counts
datasets where HiPPO-FM is higher, equal, or lower.}
\label{tab:uea-stats}
\input{appendix/uea_stats_table}
\end{table}

\subsection{Event benchmark details}
\label{app:event-details}
\input{appendix/event_details}

\subsection{PhysioNet extrapolation results}
\label{app:cru-results}
\input{appendix/results}

%% file: appendix/historical_spiral.tex
\begingroup
\setlength{\abovedisplayskip}{2pt plus 1pt minus 1pt}
\setlength{\belowdisplayskip}{3pt plus 1pt minus 1pt}
\setlength{\abovedisplayshortskip}{1pt plus 1pt minus 1pt}
\setlength{\belowdisplayshortskip}{2pt plus 1pt minus 1pt}
\setlength{\jot}{2pt}

This subsection provides the generator equations and a qualitative companion
to the main-text Spiral comparison. The visualization uses archived
5,000-iteration checkpoints; the shared data split, seeds, and main setting
are defined in Appendix~\ref{app:spiral-protocol}. The additional $\rho$ values
are a robustness check and are already summarized in the main table. The
archive is used only for this qualitative figure and does not change the
main-table training protocol.

For normalized time $t\in[s,e]$, the counter-clockwise family is
\begin{align}
x_{\rm cc}(t) &= (a+bt)\cos(t)+5,\\
y_{\rm cc}(t) &= (a+bt)\sin(t)+5.
\end{align}
The clockwise family reverses the radial-angular coordinate. With
$z(t)=e+1-t$ and $r(t)=a+50b/z(t)$, we use
\begin{align}
x_{\rm cw}(t) &= r(t)\cos(z(t))-5,\\
y_{\rm cw}(t) &= r(t)\sin(z(t))-5.
\end{align}
Clean trajectories are generated first, followed by Gaussian observation noise
on the sparse input coordinates; clean trajectories remain the evaluation
targets. Interpolation is measured on the observed half and extrapolation on
the withheld half. Numerical aggregates remain in the main table.

For the qualitative comparison, the baselines use their respective temporal
or continuous-time readouts, while HiPPO-FM uses the functional readout
described in Section~\ref{sec:method}.

\begin{figure}[H]
\centering
\includegraphics[width=\linewidth]{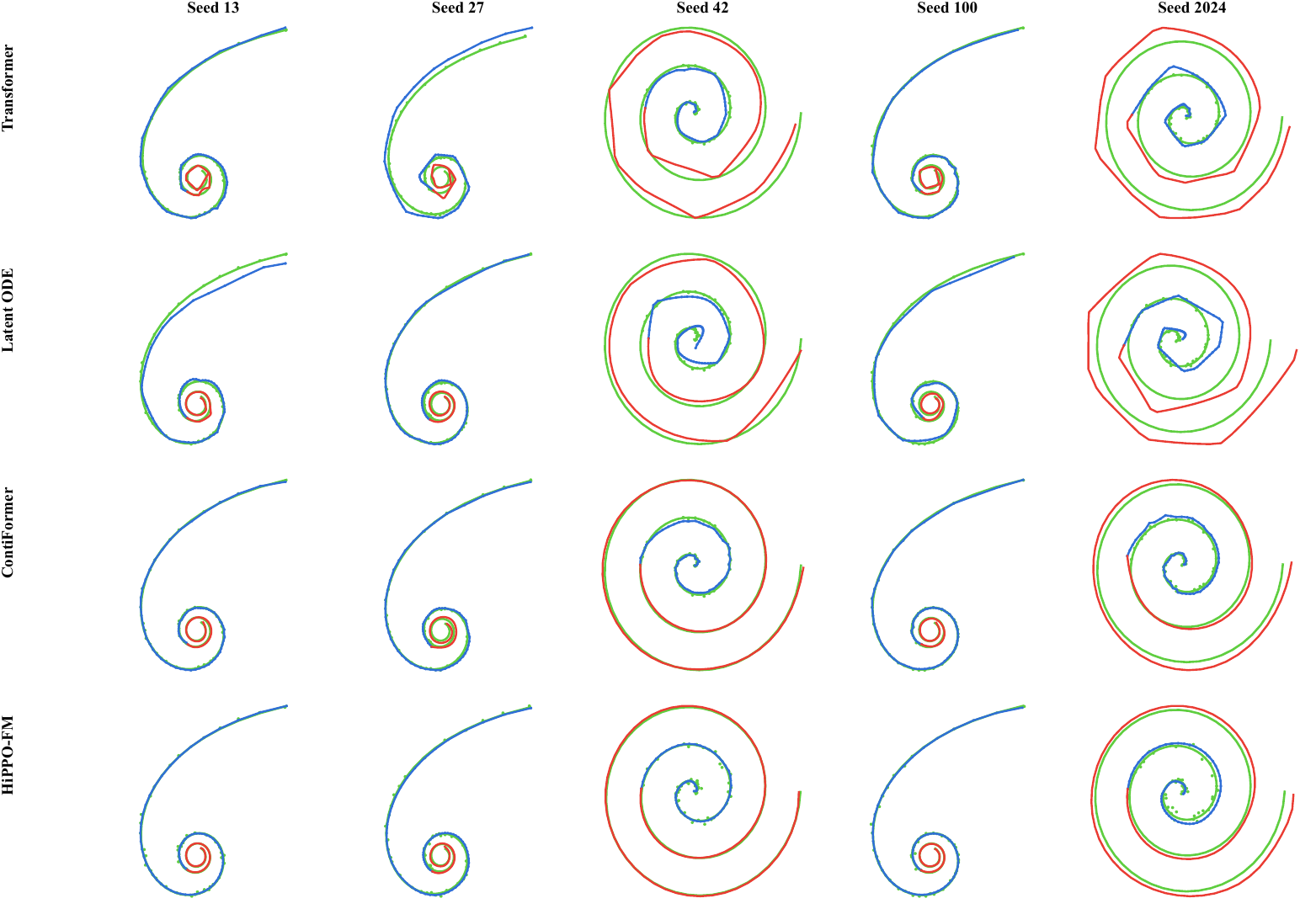}
\caption{\textbf{Five-seed qualitative Spiral visualizations under the
stated experimental setting.} Columns are seeds $\{13,27,42,100,2024\}$ and all rows
use the same plotting routine with 50 observed support points. Rows show
Transformer, Latent ODE, ContiFormer, and HiPPO-FM. This visualization is
baseline-aligned and qualitative; the main-text aggregate table reports the
$N_{\rm obs}=30$ setting.}
\label{fig:spiral-appendix-grid}
\end{figure}
\endgroup

%% file: appendix/uea_details.tex
The following tables expand Table~\ref{tab:uea} to mean test accuracy for all
20 datasets and three observation-drop settings. Row-wise boldface marks the
highest available accuracy. ACC is the macro average over reported cells, and
Rank is computed per dataset over the displayed methods; t-PatchGNN contributes
only where a cell is reported. Earlier MHA-to-FM settings, Parallel, V4,
pooling alternatives, and external adaptations are excluded. The Overall
columns in the main text average the three macro-ACC columns with equal weight;
the per-dataset Rank rows below are retained for transparency.

\begin{table}[H]
\centering
\scriptsize
\caption{UEA test accuracy with 30\% of observations dropped under the
fixed early-stopping procedure. Entries are three-seed means; $\dagger$ marks
the declared baseline-reporting exception described in the text.}
\label{tab:uea-drop30}
\setlength{\tabcolsep}{2.1pt}
\resizebox{\linewidth}{!}{%
\begin{tabular}{@{}lrrrrrrrrrrr@{}}
\toprule
Dataset & GRU-D & GRU-$\Delta t$ & ODE-RNN & CADN & N. CDE & S5 & mTAN & TST & ContiFormer & t-PatchGNN & HiPPO-FM\\
\midrule
AWR & 0.8467 & 0.8878 & 0.8878 & 0.9378 & 0.7056 & 0.8933 & 0.7322 & 0.9244 & 0.9322 & 0.7789 & \textbf{0.9500}\\
BM & 0.8417 & 0.8917 & 0.8417 & 0.8500 & 0.3500 & 0.8167 & 0.7500 & \textbf{0.9583} & 0.9250 & 0.9500 & 0.9000\\
CT & 0.9740 & 0.9705 & 0.9817 & 0.9612 & 0.8572 & \textbf{0.9749} & 0.4053 & 0.9640 & 0.9687 & 0.9615 & 0.9626\\
DDG & 0.5067 & 0.5600 & 0.5600 & 0.5267 & 0.1933 & 0.5733 & 0.6067 & 0.6600 & 0.6600 & -- & \textbf{0.6667}\\
EP & 0.9203 & 0.9251 & 0.9155 & 0.9010 & 0.4662 & 0.8696 & 0.5894 & 0.9469 & 0.9058 & 0.9251 & \textbf{0.9565}\\
ER & 0.7679 & 0.7568 & 0.7309 & 0.8259 & 0.7259 & 0.8173 & 0.7136 & 0.8321 & \textbf{0.8358} & 0.7889 & 0.7469\\
FM & 0.5500 & 0.5167 & 0.5167 & 0.5233 & 0.5233 & 0.5467 & \textbf{0.5600} & 0.5200 & 0.4500 & 0.5367 & 0.5433\\
HMD & 0.2973 & 0.3153 & 0.2928 & 0.2432 & 0.2027$^{\dagger}$ & \textbf{0.4324} & 0.2838 & 0.3288 & 0.3018 & 0.2973 & 0.3694\\
HW & 0.2055 & 0.2545 & 0.1580 & \textbf{0.2588} & 0.0886 & 0.2561 & 0.0749 & 0.1976 & 0.1549 & 0.1576 & 0.1447\\
HB & 0.6683 & 0.6634 & 0.6504 & \textbf{0.6959} & 0.5447 & 0.6098 & 0.6829 & 0.6553 & 0.6715 & 0.6911 & 0.6537\\
JV & 0.9685 & 0.9694 & 0.9667 & 0.9613 & 0.9685 & 0.9595 & 0.9550 & \textbf{0.9784} & 0.9766 & 0.8766 & 0.9631\\
LB & 0.7685 & 0.7870 & 0.7926 & 0.8593 & \textbf{0.8704} & 0.7778 & 0.2889 & 0.8056 & 0.8093 & 0.5870 & 0.7833\\
LSST & 0.4692 & 0.4695 & 0.5088 & 0.4640 & 0.0933 & 0.5377 & 0.4501 & 0.4938$^{\dagger}$ & \textbf{0.5969} & -- & 0.5603\\
NATOPS & 0.9278 & \textbf{0.9444} & 0.9389 & 0.9259 & 0.7204 & 0.9296 & 0.7204 & 0.8574 & 0.9019 & 0.7852 & 0.9130\\
PEMS & 0.6956 & 0.6513 & 0.6821 & 0.7534 & 0.6994 & 0.7380 & 0.7572 & 0.7726 & 0.7842 & -- & \textbf{0.8054}\\
PD & 0.8526 & 0.8095 & 0.8433 & 0.8435 & 0.3878 & 0.9301 & 0.4528 & 0.7925 & 0.6518 & 0.9066 & \textbf{0.9302}\\
RS & 0.7390 & 0.7500 & 0.7763 & \textbf{0.7785} & 0.3311 & 0.7018 & 0.4890 & 0.7259 & 0.6711 & 0.7215 & 0.7193\\
SR & 0.6018 & 0.6553 & 0.6268 & 0.6098 & 0.4881 & \textbf{0.8578} & 0.6644 & 0.7452 & 0.6985 & 0.7861 & 0.7418\\
SA & 0.9615 & 0.9554 & 0.9662 & 0.9375 & 0.3232 & \textbf{0.9864} & 0.4869 & 0.9422 & 0.9507 & 0.9565 & 0.9536\\
UGL & 0.7167 & 0.7760 & 0.7854 & \textbf{0.8333} & 0.7729 & 0.7844 & 0.5104 & 0.6552 & 0.6667 & 0.7271 & 0.7281\\
\midrule
ACC $\uparrow$ & 0.7140 & 0.7255 & 0.7211 & 0.7345 & 0.5156 & \textbf{0.7497} & 0.5587 & 0.7378 & 0.7257 & 0.7314 & 0.7496\\
Rank $\downarrow$ & 5.88 & 5.35 & 5.75 & 5.38 & 9.18 & \underline{4.70} & 8.58 & 4.78 & 5.18 & 5.88 & \textbf{4.60}\\
\bottomrule
\end{tabular}}
\end{table}

\begin{table}[H]
\centering
\scriptsize
\caption{UEA test accuracy with 50\% of observations dropped under the
fixed early-stopping procedure. Entries are three-seed means; $\dagger$ marks
the declared baseline-reporting exception described in the text.}
\label{tab:uea-drop50}
\setlength{\tabcolsep}{2.1pt}
\resizebox{\linewidth}{!}{%
\begin{tabular}{@{}lrrrrrrrrrrr@{}}
\toprule
Dataset & GRU-D & GRU-$\Delta t$ & ODE-RNN & CADN & N. CDE & S5 & mTAN & TST & ContiFormer & t-PatchGNN & HiPPO-FM\\
\midrule
AWR & 0.8611 & 0.8844 & 0.8900 & 0.9367 & 0.6933 & 0.8911 & 0.7189 & 0.9233 & 0.9367 & 0.7756 & \textbf{0.9544}\\
BM & 0.9000 & 0.9333 & 0.9250 & 0.9083 & 0.3083 & 0.7750 & 0.8500 & 0.9417 & 0.9417 & \textbf{0.9750} & 0.9667\\
CT & 0.9656 & 0.9754 & 0.9779 & 0.9181 & 0.8647 & \textbf{0.9659} & 0.3942 & 0.9582 & 0.9663 & 0.9568 & 0.9582\\
DDG & 0.5667 & 0.5533 & 0.5600 & 0.5467 & 0.2267 & 0.5467 & 0.5867 & 0.6200 & 0.6600 & -- & 0.6467\\
EP & 0.9348 & 0.9493 & 0.9275 & 0.9251 & 0.4541 & 0.8575 & 0.5628 & 0.9517 & 0.9130 & 0.9155 & \textbf{0.9614}\\
ER & 0.7827 & 0.7728 & 0.7469 & 0.8284 & 0.7346 & 0.7988 & 0.7123 & 0.8321 & \textbf{0.8420} & 0.7556 & 0.7247\\
FM & 0.5200 & 0.5000 & 0.4867 & \textbf{0.5300} & 0.5067 & 0.5167 & 0.5033 & 0.5200 & 0.5067 & 0.4967 & 0.5267\\
HMD & 0.2838 & 0.2838 & 0.3108 & 0.3333 & 0.2297$^{\dagger}$ & \textbf{0.4234} & 0.2793 & 0.3063 & 0.3198 & 0.3514 & 0.3423\\
HW & 0.1863 & \textbf{0.2361} & 0.1596 & 0.2161 & 0.0878 & 0.2063 & 0.0690 & 0.1616 & 0.1408 & 0.1231 & 0.1224\\
HB & 0.6488 & 0.6748 & 0.6309 & 0.6813 & 0.6098 & \textbf{0.7154} & 0.6537 & 0.6585 & 0.6683 & 0.6634 & 0.6537\\
JV & 0.9532 & 0.9550 & 0.9622 & 0.9423 & 0.9604 & \textbf{0.9658} & 0.9432 & 0.9577 & 0.9559 & 0.6306 & 0.9523\\
LB & 0.7352 & 0.7556 & 0.7556 & 0.8167 & \textbf{0.8630} & 0.7463 & 0.2704 & 0.7926 & 0.8111 & 0.5407 & 0.7944\\
LSST & 0.4566 & 0.4669 & 0.4669 & 0.4393 & 0.1160 & 0.5188 & 0.4324 & 0.4620$^{\dagger}$ & \textbf{0.5673} & -- & 0.5270\\
NATOPS & \textbf{0.9370} & 0.9167 & \textbf{0.9370} & 0.9204 & 0.8093 & 0.8926 & 0.6778 & 0.8333 & 0.9296 & 0.7778 & 0.9185\\
PEMS & 0.6782 & 0.6455 & 0.7013 & 0.7399 & 0.6744 & 0.6532 & 0.7206 & 0.7726 & \textbf{0.8112} & -- & 0.7861\\
PD & 0.7707 & 0.6662 & 0.7216 & 0.6275 & 0.2249 & 0.7616 & 0.3630 & 0.5171 & 0.5587 & \textbf{0.7985} & 0.7906\\
RS & 0.7346 & 0.7346 & 0.7456 & \textbf{0.7632} & 0.2785 & 0.7127 & 0.5000 & 0.7259 & 0.6294 & 0.6447 & 0.7325\\
SR & 0.5404 & 0.6291 & 0.6200 & 0.6837 & 0.4972 & \textbf{0.8794} & 0.6519 & 0.7486 & 0.6849 & 0.7907 & 0.7315\\
SA & 0.9488 & 0.9451 & 0.9586 & 0.9322 & 0.4390 & \textbf{0.9760} & 0.4846 & 0.9280 & 0.9579 & 0.9530 & 0.9468\\
UGL & 0.7552 & 0.7698 & \textbf{0.8167} & 0.8125 & 0.7698 & 0.8115 & 0.5198 & 0.6479 & 0.6677 & 0.7115 & 0.6917\\
\midrule
ACC $\uparrow$ & 0.7080 & 0.7124 & 0.7150 & 0.7251 & 0.5174 & 0.7307 & 0.5447 & 0.7130 & 0.7235 & 0.6977 & \textbf{0.7364}\\
Rank $\downarrow$ & 5.95 & 5.68 & 5.23 & 4.85 & 9.00 & 4.73 & 9.23 & 5.28 & \textbf{4.43} & 6.47 & \underline{4.50}\\
\bottomrule
\end{tabular}}
\end{table}

\begin{table}[H]
\centering
\scriptsize
\caption{UEA test accuracy with 70\% of observations dropped under the
fixed early-stopping procedure. Entries are three-seed means; $\dagger$ marks
the declared baseline-reporting exception described in the text.}
\label{tab:uea-drop70}
\setlength{\tabcolsep}{2.1pt}
\resizebox{\linewidth}{!}{%
\begin{tabular}{@{}lrrrrrrrrrrr@{}}
\toprule
Dataset & GRU-D & GRU-$\Delta t$ & ODE-RNN & CADN & N. CDE & S5 & mTAN & TST & ContiFormer & t-PatchGNN & HiPPO-FM\\
\midrule
AWR & 0.8611 & 0.8944 & 0.9067 & \textbf{0.9356} & 0.7778 & 0.8956 & 0.7600 & 0.9133 & 0.9144 & 0.7600 & 0.9322\\
BM & 0.8833 & 0.8833 & 0.9333 & 0.8750 & 0.2667 & 0.7917 & 0.8250 & 0.9333 & 0.9333 & \textbf{0.9917} & 0.9500\\
CT & 0.9615 & 0.9731 & 0.9659 & 0.8728 & 0.8635 & \textbf{0.9547} & 0.3679 & 0.9420 & 0.9633 & 0.9422 & 0.9325\\
DDG & 0.5933 & 0.5467 & 0.5200 & 0.5400 & 0.2467 & 0.5667 & 0.5733 & 0.5800 & 0.5733 & -- & \textbf{0.6867}\\
EP & 0.9058 & 0.9348 & 0.9251 & 0.8865 & 0.5531 & 0.8623 & 0.5362 & 0.9275 & 0.9082 & 0.9034 & \textbf{0.9565}\\
ER & 0.7556 & 0.7457 & 0.7111 & 0.7691 & 0.7765 & 0.7654 & 0.6543 & 0.7741 & \textbf{0.8333} & 0.7395 & 0.6210\\
FM & 0.4900 & \textbf{0.5533} & 0.5200 & 0.5200 & 0.5033 & \textbf{0.5533} & 0.4533 & 0.4600 & 0.4867 & 0.5333 & 0.4767\\
HMD & 0.2838 & 0.2658 & 0.3514 & 0.2928 & 0.2658 & \textbf{0.3919} & 0.3018 & 0.2658 & 0.2793 & 0.3108 & 0.3559\\
HW & 0.1725 & \textbf{0.1867} & 0.1502 & 0.1376 & 0.0796 & 0.1427 & 0.0537 & 0.1302 & 0.1220 & 0.0996 & 0.1024\\
HB & 0.6472 & 0.6846 & 0.6439 & 0.6520 & 0.6163 & 0.6325 & \textbf{0.6959} & 0.6179 & 0.6911 & 0.6569 & 0.6634\\
JV & 0.9243 & 0.9333 & 0.9297 & 0.9423 & 0.9288 & 0.9414 & 0.9324 & \textbf{0.9495} & \textbf{0.9495} & 0.7297 & 0.9468\\
LB & 0.5537 & 0.6444 & 0.6611 & 0.7407 & \textbf{0.8037} & 0.6426 & 0.2222 & 0.7796 & 0.7519 & 0.6741 & 0.7056\\
LSST & 0.4303 & 0.4565 & 0.4244 & 0.4347 & 0.1245 & 0.4808 & 0.4207 & 0.4556$^{\dagger}$ & \textbf{0.5242} & 0.4109 & 0.4861\\
NATOPS & 0.9074 & \textbf{0.9222} & 0.8981 & 0.9111 & 0.7481 & 0.9056 & 0.6167 & 0.8611 & 0.8981 & 0.7222 & 0.9148\\
PEMS & 0.6455 & 0.6185 & 0.6724 & 0.6686 & 0.4297 & 0.6474 & 0.6647 & \textbf{0.7553} & 0.7476 & -- & 0.7148\\
PD & 0.6142 & 0.5006 & 0.5880 & 0.4982 & 0.1454 & 0.6118 & 0.2933 & 0.4018 & 0.4751 & \textbf{0.6809} & 0.6561\\
RS & 0.6711 & 0.6798 & 0.7171 & \textbf{0.7346} & 0.2851 & 0.6075 & 0.4430 & 0.6908 & 0.6162 & 0.5636 & 0.6316\\
SR & 0.6303 & 0.6246 & 0.6735 & 0.5825 & 0.5358 & \textbf{0.8635} & 0.6553 & 0.7406 & 0.7099 & 0.8100 & 0.7804\\
SA & 0.9218 & 0.9113 & 0.9335 & 0.9110 & 0.3352 & \textbf{0.9573} & 0.4705 & 0.8739 & 0.9297 & 0.9163 & 0.9053\\
UGL & 0.7323 & 0.7917 & 0.7729 & \textbf{0.8125} & 0.7698 & 0.8094 & 0.4979 & 0.6625 & 0.7125 & 0.7156 & 0.6969\\
\midrule
ACC $\uparrow$ & 0.6793 & 0.6876 & 0.6949 & 0.6859 & 0.5028 & 0.7012 & 0.5219 & 0.6857 & 0.7010 & 0.6756 & \textbf{0.7058}\\
Rank $\downarrow$ & 6.03 & 5.10 & 5.20 & 5.38 & 8.90 & 5.08 & 8.80 & 5.63 & \textbf{4.53} & 6.36 & \underline{4.55}\\
\bottomrule
\end{tabular}}
\end{table}

%% file: appendix/uea_stats_table.tex
\begin{tabular*}{\linewidth}{@{\extracolsep{\fill}}llcccc@{}}
\toprule
Drop & Baseline & $\Delta$ACC (pp) & W/T/L & $p$ & $p_{\mathrm{Holm}}$\\
\midrule
30\% & S5 & $-0.01$ & 11/0/9 & 0.869 & 1.00\\
30\% & ContiFormer & $+2.39$ & 12/0/8 & 0.216 & 1.00\\
50\% & S5 & $+0.57$ & 11/0/9 & 0.812 & 1.00\\
50\% & ContiFormer & $+1.30$ & 9/0/11 & 0.455 & 1.00\\
70\% & S5 & $+0.46$ & 12/0/8 & 0.756 & 1.00\\
70\% & ContiFormer & $+0.48$ & 9/0/11 & 0.970 & 1.00\\
\bottomrule
\end{tabular*}

%% file: appendix/event_details.tex
The event benchmark contains one synthetic and five real-world marked
sequences. The reported dataset statistics are: Synthetic ($K=10$, 8,618 events,
average length 17, 375/125 train/test sequences), Neonate ($K=1$, 534
events, average length 11, 38/10 sequences), Traffic ($K=2$, 154,747
events, average length 60, 1,938/647 sequences), MIMIC ($K=75$, 2,419
events, average length 4, 527/65 sequences), StackOverflow ($K=22$,
480,413 events, average length 72, 4,777/1,326 sequences), and BookOrder
($K=2$, 414,800 events, average length 2,074, 90/100 sequences). Here $K$
denotes the number of event types and the two sequence counts are train/test
counts.

The synthetic benchmark uses ten event types and 500 generated sequences
with four-fold cross-validation. Neonate records seizure-start events from
79 patients; only 48 patients have at least two seizures. Traffic contains
two event types derived from large changes in hourly sensor measurements,
with events close in time deduplicated. MIMIC treats each ICU visit as a
sequence. BookOrder contains buy and sell events with sequences clipped to
length 500 and a maximum inter-event gap of 70, while StackOverflow records
user badge-reward histories over two years. The real-data folds follow the
partition counts above; MIMIC, StackOverflow, and BookOrder use five folds,
whereas Neonate and Traffic use four folds.

%% file: appendix/results.tex
Table~\ref{tab:cru} gives the complete five-seed record for the PhysioNet
comparison in Section~\ref{sec:physionet}. It reports hidden-region MSE,
seconds per epoch, peak allocated memory, and parameter count; values are mean
$\pm$ sample standard deviation where applicable. The protocol is given in Appendix~\ref{app:physionet-protocol}.

\begin{table}[h]
\centering\small
\caption{\textbf{PhysioNet extrapolation, five seeds.} MSE and seconds per epoch are mean $\pm$ sample SD. Memory is peak allocated memory.}
\label{tab:cru}
\setlength{\tabcolsep}{4pt}
\begin{tabular}{@{}llrrrr@{}}
\toprule
Dataset & Model & MSE $\downarrow$ & s/epoch $\downarrow$ & GiB $\downarrow$ & Parameters\\
\midrule
PhysioNet & CRU & $0.006465\smallpm0.000409$ & $45.84\smallpm12.13$ & 4.121 & 28,654\\
& HiPPO-FM & $\mathbf{0.005099}\smallpm0.000100$ & $\mathbf{2.01\smallpm0.14}$ & \textbf{0.196} & 55,978\\
\bottomrule
\end{tabular}
\end{table}

%% file: appendix/diagnostics.tex
\section{Efficiency, Ablation, and Memory-Rank Diagnostics}
\label{app:diagnostics}
\paragraph{Protocol.}
The diagnostics of Section~\ref{sec:efficiency} use the Spiral protocol of Table~\ref{tab:spiral-main} (Appendix~\ref{app:spiral-protocol}) at $\rho=0.02$. Each seed generates 300 trajectories with 150 time positions, noise scale 0.1, and $\rho=0.02$; 200 trajectories are used for training and 100 for testing, and an independent 50-trajectory validation set generated with seed $s+10^6$ is used for checkpoint selection by validation MAE. Each training step samples 30 observation times from the first half and a batch of 64 trajectories. All models use Adam with learning rate $10^{-2}$, gradient clipping at norm 1, and fp32; ablation and rank runs train for 5,000 iterations, and the wall-clock runs train until the time budget is exhausted. In this protocol the HiPPO-FM model (one MHA-to-FM encoder block, width 16, four heads, $M=16$) receives the observations linearly interpolated onto the evaluation grid before the encoder, and the ablation variants modify only the encoder layer. The five seeds are $\{13,27,42,100,2024\}$, and all comparisons are paired by seed.

\paragraph{Ablation variants.}
\emph{w/o encoder FM} removes the functional-memory operator from the encoder layer and keeps the scaled observed-support self-attention, layer normalization, and feed-forward network. \emph{Cosine basis} replaces the shifted-Legendre HiPPO basis with orthonormal cosine modes. \emph{No Gram solve} reads the normalized memory directly and skips the regularized Gram solve. \emph{Point features (no kernel)} writes and reads with the basis evaluated at the coordinates instead of kernel-smoothed features and keeps the finite-basis Gram solve. The Full row is the clean full-model run with the same seeds and is shared by all retained variants.

\input{appendix/diagnostics_tables}

\paragraph{Wall-clock budget.}
The 30\,s budget is checked once per iteration, so the final iteration can exceed it; ContiFormer completes nine iterations in 32.3\,s. Every model uses the same data pipeline, optimizer, and validation schedule; baselines keep their default hyperparameters. The comparison therefore characterizes optimization under a fixed implementation and hardware budget; Table~\ref{tab:complexity} gives the compact asymptotic summary.

\paragraph{Retained-state diagnostic.}
Table~\ref{tab:efficiency-full} uses frozen matched-decoder checkpoints in which the same trained FM encoder is read either by the FM head or by a token cross-attention decoder. Timed regions exclude data transfer and are synchronized before and after each call. The persistent state counts unique tensor storage retained after encoding for a batch of 20 histories. For the FM head this is the solved coefficient tensor and the coordinate bounds (82.1\,kB, independent of $N$) plus the $N$-length Boolean validity mask, which the readout ignores; the mask accounts for the growth from 82.7\,kB to 102.6\,kB. The token cache grows linearly in $N$. The token decoder is faster per call, and peak transient allocation is similar for both readouts.

\begin{figure}[H]
\centering
\includegraphics[width=\linewidth]{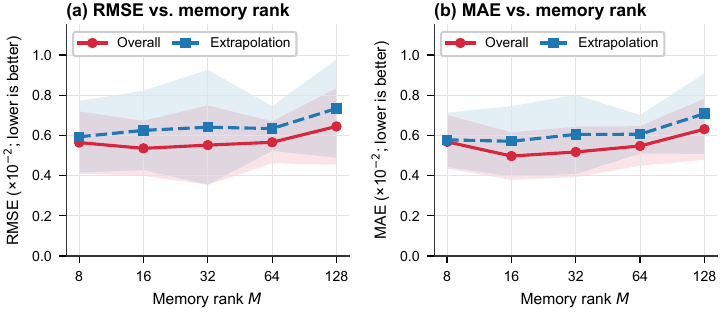}
\caption{Memory-rank sensitivity diagnostic on Spiral (five seeds per rank; lines are means and bands are $\pm1$ sample SD). \textbf{(a)} RMSE and \textbf{(b)} MAE on all evaluation coordinates (Overall) and on the withheld half (Extrapolation). The sweep is descriptive under the fixed training protocol and does not imply a monotonic rank--accuracy relation.}
\label{fig:rank-scaling}
\end{figure}

\paragraph{Memory-rank sensitivity diagnostic.}
Figure~\ref{fig:rank-scaling} trains the full model with $M\in\{8,16,32,64,128\}$ under the same protocol; $M=16$ reuses the Full runs above. Overall RMSE is lowest at $M=16$, extrapolation RMSE is lowest at $M=8$, and $M=128$ is worst on both, with overlapping seed variation for $M\le64$. We treat this sweep as a sensitivity diagnostic under the fixed protocol, rather than as evidence that rank alone determines error or that larger ranks are uniformly better.

%% file: appendix/diagnostics_tables.tex
\begin{table}[!t]
\centering
\caption{Spiral ablation variants ($\times10^{-2}$, mean $\pm$ sample SD over five paired seeds). $k/5$ counts seeds with higher overall RMSE than Full; $p$-values are two-sided paired $t$ and exact Wilcoxon tests on overall RMSE ($n=5$, minimum attainable Wilcoxon $p=0.0625$).}
\label{tab:ablation-full}
\scriptsize
\setlength{\tabcolsep}{4.2pt}
\begin{tabular*}{\linewidth}{@{\extracolsep{\fill}}lcccccc@{}}
\toprule
Variant & Overall RMSE & Extrap.\ RMSE & $k/5$ & $t$-test $p$ & Wilcoxon $p$ & Params\\
\midrule
Full & $0.54\smallpm0.14$ & $0.63\smallpm0.20$ & -- & -- & -- & 4037\\
\midrule
w/o encoder FM (MHA only) & $9.88\smallpm0.59$ & $13.79\smallpm0.77$ & 5/5 & $3.5\times10^{-6}$ & 0.0625 & 3363\\
Cosine basis & $0.61\smallpm0.10$ & $0.66\smallpm0.15$ & 4/5 & $0.24$ & 0.3125 & 4037\\
No Gram solve & $0.63\smallpm0.07$ & $0.64\smallpm0.07$ & 4/5 & $0.10$ & 0.125 & 4037\\
Point features (no kernel) & $0.49\smallpm0.06$ & $0.50\smallpm0.05$ & 3/5 & $0.56$ & 1 & 4037\\
\bottomrule
\end{tabular*}
\end{table}

\begin{table}[!t]
\centering
\caption{Equal wall-clock training on Spiral (nominal 30\,s, one RTX~5090, five paired seeds). Test errors are $\times10^{-2}$; elapsed time includes data preparation and validation every 25 iterations; quality is validation MAE $\le0.02$; peak memory is the maximum allocated GPU memory (MB).}
\label{tab:wallclock-full}
\scriptsize
\setlength{\tabcolsep}{3pt}
\begin{tabular*}{\linewidth}{@{\extracolsep{\fill}}lcccccccc@{}}
\toprule
Model & Test RMSE & Test MAE & Iterations & Time (s) & Reached & Time to quality (s) & Peak MB & Params\\
\midrule
HiPPO-FM & $0.483\smallpm0.028$ & $0.499\smallpm0.026$ & $6507\smallpm241$ & $30.01$ & 5/5 & $3.23\smallpm0.56$ & 116 & 4037\\
Transformer & $1.015\smallpm0.053$ & $0.949\smallpm0.083$ & $18469\smallpm233$ & $30.00$ & 5/5 & $3.43\smallpm0.30$ & 59 & 4530\\
Neural ODE & $8.924\smallpm1.127$ & $9.018\smallpm1.074$ & $231\smallpm21$ & $30.32$ & 0/5 & not reached & 31 & 1186\\
Latent ODE & $9.415\smallpm1.317$ & $9.035\smallpm1.374$ & $183\smallpm17$ & $30.37$ & 0/5 & not reached & 32 & 1322\\
Neural CDE & $16.890\smallpm7.698$ & $17.851\smallpm8.542$ & $87\smallpm0$ & $30.28$ & 0/5 & not reached & 22 & 4306\\
ContiFormer & $24.841\smallpm6.562$ & $28.735\smallpm8.403$ & $9\smallpm0$ & $32.26$ & 0/5 & not reached & 3961 & 3938\\
\bottomrule
\end{tabular*}
\end{table}

\begin{table}[!t]
\centering
\caption{Fixed-checkpoint observation scaling (batch of 20 histories, 50 warm-up and 200 timed repetitions, CUDA-synchronized medians). FM: HiPPO-FM state; Token: the same trained encoder read by a token cross-attention decoder that caches the encoded observations. State in kB, latency in ms, peak allocation in MB.}
\label{tab:efficiency-full}
\scriptsize
\setlength{\tabcolsep}{4.2pt}
\begin{tabular*}{\linewidth}{@{\extracolsep{\fill}}rccccccccc@{}}
\toprule
& \multicolumn{3}{c}{Persistent state} & \multicolumn{2}{c}{Encoding} & \multicolumn{2}{c}{Readout} & \multicolumn{2}{c}{Peak alloc.}\\
\cmidrule(lr){2-4}\cmidrule(lr){5-6}\cmidrule(lr){7-8}\cmidrule(lr){9-10}
$N$ & FM & Token & Ratio & FM & Token & FM & Token & FM & Token\\
\midrule
32 & 82.7 & 328.5 & 4.0 & 8.75 & 5.83 & 1.60 & 0.65 & 22 & 23\\
64 & 83.4 & 656.8 & 7.9 & 8.79 & 5.82 & 1.62 & 0.66 & 24 & 26\\
128 & 84.6 & 1313.4 & 15.5 & 8.86 & 5.85 & 1.62 & 0.66 & 33 & 34\\
256 & 87.2 & 2626.7 & 30.1 & 8.82 & 5.86 & 1.61 & 0.66 & 68 & 69\\
512 & 92.3 & 5253.3 & 56.9 & 8.85 & 5.89 & 1.62 & 0.66 & 199 & 202\\
1024 & 102.6 & 10506.4 & 102.4 & 8.90 & 6.09 & 1.63 & 0.85 & 713 & 719\\
\bottomrule
\end{tabular*}
\end{table}